\documentclass[acmnone]{acmart}
\usepackage{booktabs} % For formal tables
\usepackage{xcolor}
\usepackage[normalem]{ulem}

\colorlet{yellowish}{green!10!orange}
\colorlet{greenish}{green!70!orange}
\colorlet{blueish}{blue!70}

\usepackage[ruled]{algorithm2e} % For algorithms
\usepackage[noabbrev, capitalise]{cleveref}
\usepackage{caption}
\usepackage{subcaption}
\usepackage{dirtytalk}
\usepackage{selectp}
\usepackage{booktabs}
\usepackage{multirow}
\usepackage{fontawesome5}
\usepackage{mathtools}
\usepackage[normalem]{ulem}

\SetAlFnt{\small}
\SetAlCapFnt{\small}
\SetAlCapNameFnt{\small}
\SetAlCapHSkip{0pt}

\newcommand{\ourmethod}{\textit{PlaMo}}

\acmJournal{TOG}
\begin{document}
\title{PlaMo: Plan and Move in Rich 3D Physical Environments}

% DO NOT ENTER AUTHOR INFORMATION FOR ANONYMOUS TECHNICAL PAPER SUBMISSIONS TO SIGGRAPH 2019!
% TODO: Fill in everyone

\author{Assaf Hallak}
\authornote{Both authors contributed equally to this research.}
\author{Gal Dalal\textsuperscript{*}}

\affiliation{%
  \institution{NVIDIA Research}
   \country{Israel}
}
\email{ahallak@nvidia.com, gdalal@nvidia.com}

\author{Chen Tessler}
\affiliation{%
  \institution{NVIDIA Research}
  \country{Israel}
}
\author{Kelly Guo}
\affiliation{%
  \institution{NVIDIA Research}
  \country{USA}
}
\author{Shie Mannor}
\affiliation{%
  \institution{NVIDIA Research}
  \country{Israel}
}
\author{Gal Chechik}
\affiliation{%
  \institution{NVIDIA Research}
  \country{Israel}
}

\setcopyright{none}
\fancyfoot[L]{Preprint}
% Add your custom footer text if needed
\thispagestyle{plain} % Apply plain page style to the first page

\begin{abstract}

    Controlling humanoids in complex physically simulated worlds is a long-standing challenge with numerous applications in gaming, simulation, and visual content creation. In our setup, given a rich and complex 3D scene, the user provides a list of instructions composed of target locations and locomotion types. To solve this task we present PlaMo, a scene-aware path planner and a robust physics-based controller. The path planner produces a sequence of motion paths, considering the various limitations the scene imposes on the motion, such as location, height, and speed. Complementing the planner, our control policy generates rich and realistic physical motion adhering to the plan. We demonstrate how the combination of both modules enables traversing complex landscapes in diverse forms while responding to real-time changes in the environment. \\
  \textbf{Video:} {\color{blue}\url{https://youtu.be/wWlqSQlRZ9M}}.
\end{abstract}





\begin{teaserfigure}
  \includegraphics[width=\textwidth]{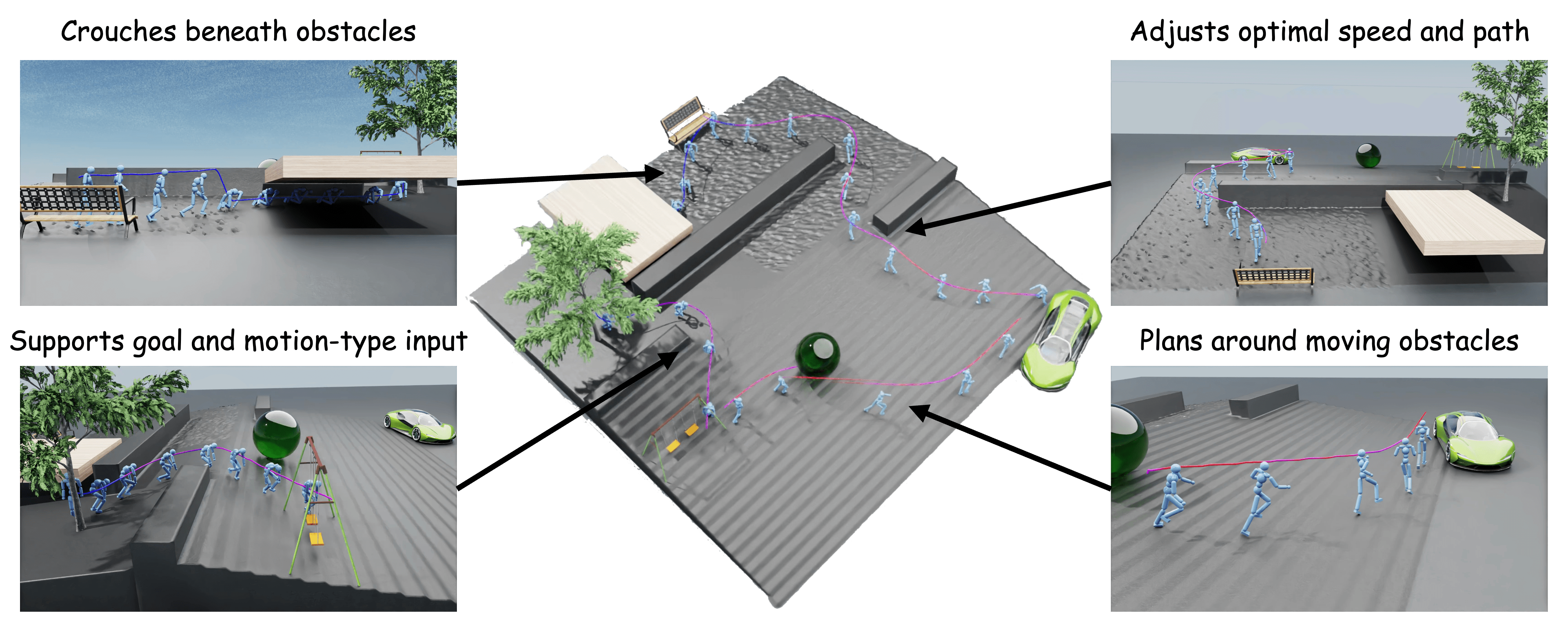}
  \caption{\textbf{Path planning and motion control} \ourmethod{} animates a humanoid in a complex 3D scene. The input consists of a physically simulated 3D scene and a series of text instructions describing high-level navigation landmarks and locomotion type ("Crouch-walk from the tree to the swing"). The scene may contain diverse terrain (gravel, stairs), 3D obstacles, and dynamic obstacles (here, a green ball).  The output is a sequence of motor actuations controlling a humanoid character.  \ourmethod{} produces a planned path together with a head-height and speed profile that matches the textual guidance (locomotion type) and the constraints of the movement controller. {A. The controller adapts to the 3D environment, producing a crawl locomotion under the obstacle. B. Supports various motion types. C. continuously re-planning the path allows to avoid moving obstacles. D. Locomotion speed is adjusted depending on terrain and obstacles. } }
  \label{fig:teaser}
\end{teaserfigure}

\maketitle

Physics-consistent animation of human characters is a challenging task with numerous applications, from automating game design through simulating factories and urban areas to training autonomous vehicles in pedestrian scenarios. In some cases, one may want to control the animation using text instructions that capture high-level goals, like "Run to the road once the car passes by" or "Hide behind the large tree". These tasks involve two types of control problems: First, planning a sequence of motions for reaching the goal with a natural behavior; Then, controlling the actual motion of the character to generate a natural-looking movement. 

Previous studies in physics-based animation have usually focused on one of these two challenges, often addressing a simplified setup. For planning, \citet{2017-TOG-deepLoco} developed a method for long-range planning of a bipedal character, and \cite{rempe2023trace} generates a distribution of trajectories based on user-defined characteristics. Other studies focused on movement control: mimicking \cite{peng2018deepmimic}, interleaving motion types \cite{tessler2023calm},  interactions with objects \cite{wang2023physhoi} or people \cite{zhang2023simulation} and crowd simulation \cite{rempe2023trace} (see related method section).

However, the two tasks are interdependent. Planning which motions to take depends on the physical characteristics of each motion and the character. Then, executing the motion depends on the sequence and longer-horizon goal. Importantly, this interdependence becomes stronger when environments become more complex. Handling 3D obstacles, uneven terrain, and dynamic objects may require the planner to override user instruction and revert to motions that are feasible in the specific environment. In this work, we address the problem of training physics-based motion models for humanoids that can handle long-horizon planning and motion control in intricate environments. 

Our approach \ourmethod{}, for \textit{Plan and Move}, combines long-horizon path planning with a novel matching motion controller. This fusion creates an agent that responds to semantic instructions from users but can adapt the plan and motion for navigating naturally through various terrains and obstacles. %, including those that are static, dynamic, or impose height restrictions. It ensures that the humanoid navigates smoothly and human-like, avoiding obstacles along the path.

When designing a path planner for this task, we need to consider several aspects. First, the planner must only produce plans that can be executed by the character's physics-based motion controller. More specifically, the character typically has a range of possible motions, such as walking, running, or crouching under an obstacle. The planner should devise plans that are within the feasibility envelope of the motion controller. Second, the capabilities of the humanoid character depend on the environment. For example, it is harder to run uphill than downhill. The planner therefore must be aware of the slope and obstacles. A similar problem occurs when turning around obstacles. Running too fast makes it harder to control a character, so the planner must match the character's velocity to the curvature of the path.

To address these challenges, we develop a planner inspired by state-of-the-art planners in the fields of robotics and autonomous driving \cite{wang2020robot,hsu2020reinforcement,alcantara2021optimal}. These techniques have proved very useful in hard real-world problems and we show here how they can be applied in physics-based character animation. Our planner operates in three stages. First, it constructs a low-resolution path using the $A^*$ algorithm. Second, it creates a high-resolution height trajectory that determines the position of a character's head along the path, taking into account 3D obstacles. 
%Clearly, this aspect has not been studied before in autonomous driving. (GalD: in fact, it does have applications for drones and robots so let's avoid determining it hasn't been done before).
Third, it sets the locomotion velocity along the path, based on the slope and curvature of the path and on the feasible movement envelope of the character, which we learn offline. This is achieved by solving a quadratic program that finds the optimal speed along the path \cite{alcantara2021optimal}. At the end of planning, it produces a sequence of head positions and velocity along the path and sends that to the motion controller. As far as we know, this is the first long-horizon planner that is aware of a complex 3D environment and character motion envelope.

Guided by the planned path, the motion controller actuates the character's joints at every time step. We condition the controller on the pose and velocity of the humanoid, surrounding terrain, and near-term path waypoints. To train the motion controller, we designed a reward function with two main components: path-following and style. The path-following component rewards the agent for successfully reaching target 3D waypoints at designated times. Our reward design considers 3D path tracking, coupled with additional terms for improving orientation and posture. For style, we use a discriminative objective \cite[AMP]{peng2021amp} encouraging natural movement. This controller is trained on random paths that take into account the correlation between height and speed, for example, crawling is typically performed slowly.

In summary, this paper makes the following novel contributions:
(1) We introduce a methodology for combining path planning and motion control within a detailed physical simulation, which we call \ourmethod{}. (2) We present an efficient method for planning long-horizon paths for animated humanoid characters, producing a feasible and safe path across uneven terrain around static, dynamic, and three-dimensional obstacles. (3) We develop a robust motion controller trained through reinforcement learning. Training to imitate human motions, the controller produces motion guided by a sequence of 3D head positions and velocities. 

\section{Related Work}

\textbf{Physics-based animation.} Prior work shows that physics simulation can be utilized for learning scene-aware controllers \cite{yuan2019ego,yuan2020residual,luo2022embodied,luo2023perpetual, won2022physics}. These controllers are robust to changes in the scene like irregular terrain, obstacles, and adversarial perturbations \cite{wang2020unicon,lee2023questenvsim,luo2023universal}. Typically, prior work focused either on complex motions in simple scenes \cite{peng2022ase, tessler2023calm, juravsky2022padl}, or simple motions within complex scenes \cite{hassan2023synthesizing, xie2020allsteps}. Closest to our work is TRACE and PACE \cite{rempe2023trace}, which focuses on pedestrian locomotion with a diffusion-based planner. To generate paths, the planning module TRACE is trained on pedestrian data, typically upright walking motions. In contrast, we focus on long-term 3D-aware navigation and locomotion. \ourmethod{} combines a short-sighted terrain-aware locomotion controller with a long-term and adaptive scene-aware path generator. The path generator plans a 3D trajectory. This plan takes into account the scene and the capabilities of the motion controller. It determines the speed and height to avoid both static and dynamic obstacles. This plan is provided to the controller. The controller then tracks the requested path along rich and complex terrains.

% \assaf{Why not trace from PACER? TRACE requires real world data of pedestrians for behavior control. In our case the obstacles have a simple motion, and the motion can be constrained (crawling, crouching), for which data does not exist.}

\textbf{Scene-aware animation.} Most prior work has used kinematic animated characters to tackle a wide range of problems such as object interaction \cite{starke2019neural, li2023object}, long-term motion generation \cite{ling2020character} and storytelling \cite{qing2023storytomotion}. However, as physics-based animation is less forgiving, prior works that use simulation have typically focused on solving specific object-interaction tasks \cite{hassan2023synthesizing}, or simple locomotion (walking or running) with short-term planning across complex terrains \cite{rempe2023trace}. We tackle the problem of long-term motion generation across complex scenes using physics-based control.

\textbf{Path planning. } There are numerous works in the world of planning motion in different levels of hierarchy. For high-level planning, the A* algorithm and its variations \cite{bell2009hyperstar, yao2010path, duchovn2014path} use dynamic programming to produce a discrete path that considers the cost and connectivity of moving across the grid. Alternatively, navigation meshes \cite{kallmann2010shortest, van2012navigation, kallmann2014navigation} use the geometric structure of the obstacle to find the shortest path. Finer control models such as Model Predictive Control (MPC; \cite{qin1997overview, holkar2010overview}) construct an optimization problem by considering the dynamics of all objects and the physical constraints of the agent while using a lookahead of a receding horizon, these models usually require an explicit knowledge on the dynamics of the agent itself. In cases where the dynamics in the scene are unknown or complicated, learning methods can be employed instead such as RL for a given simulator \cite{michels2005high, chen2022deep, zhang2023learning} or diffusion models when there is access to relevant data \cite{barquero2023belfusion, karunratanakul2023guided}. In this context, our works are most related to recent approaches for planning in autonomous driving, extending A* with additional smoothing and obstacle avoidance modules to form feasible trajectories \cite{yeong2020development, zhong2020hybrid, maw2020iada}. 

%\chen{Goal here is to motivate physics based animation. We want to say (1) prior work showed that physics based animation can learn robust reactive policies. These policies adapt to changes in the scene (object, terrain, adversarial perturbations). However, these works typically focus on complex motions in simple scenes -- ASE, CALM, PADL, PHC; or simple motions within complex scenes -- PACER, InterPhys.}\chen{Also need to talk about how these methods solve very specific tasks.}\chen{In this work we tackle long-term planning with improved motion control across complex scenes.}\chen{should probably mention that PACER takes a step in this direction, but only considers locomotion as they focus on pedestrians, however in many problems both indoors and outdoors, especially in games, it is of great interest to have 3D awareness and motion.}

%\textbf{Scene aware animation.} \chen{Here we have two lines of work. (1) kinematic models. they have tackled a very wide range of problems. GalC shared some in the WW channel, interacting with objects, long-term motion generation Scene something, etc... (2) physics based, where it is much more challenging as the problem is control. The results are much higher quality and generalize better but harder to achieve. In this work we tackle long-term motion generation across complex scenes using physics-based.}

\section{Preliminaries}
We begin with a formal introduction of the reinforcement learning (RL) setup for training our motion controller, together with the mathematical representation of our simulated scene. 
\subsection{Reinforcement Learning}
We train a goal-conditioned motion controller which we model as an RL \cite{sutton2018reinforcement} problem. In RL, an agent interacts with an environment according to a policy $\pi$. At each step $t$, the agent observes a state $s_t$ and samples an action $a_t$ from the policy $a_t \sim \pi (a_t | s_t)$. The environment then transitions to the next state $s_{t+1}$ based on the transition probability $p(s_{t+1} | s_t, a_t)$. The goal is to maximize the discounted cumulative reward, defined as
\begin{equation*}
    J = \mathbb{E}_{p(\sigma|\pi)} \left[ \sum_{t=0}^T \gamma^t r_t \middle| s_0 = s \right] \,,
\end{equation*}
where $p(\sigma|\pi) = p(s_0) \Pi_{t=0}^{T-1} p(s_{t+1}|s_t,a_t) \pi(a_t | s_t)$ is the likelihood of a trajectory $\sigma = (s_0, a_0, r_0, \ldots, s_{T-1}, a_{T-1}, r_{T-1}, s_T)$, and $\gamma \in [0, 1)$ is a discount factor that determines the effective horizon of the policy.

To find the optimal policy $\pi^*$ that maximizes the discounted cumulative reward, the popular policy gradient paradigm seeks to directly optimize the policy using the utility of complete trajectories. Perhaps the most widely used algorithm in this context is PPO \cite{schulman2017proximal}, which gradually improves the policy while maintaining stability without drifting too far from it.

% In this work, the state of the system consists of the humanoid's pose and speed, in addition to the scene information, such as the surrounding height map. As we focus on physics-based animation, the actions correspond to motor actuations for the simulated humanoid.

\begin{figure*}[ht]
    \centering
    \includegraphics[width=\linewidth]{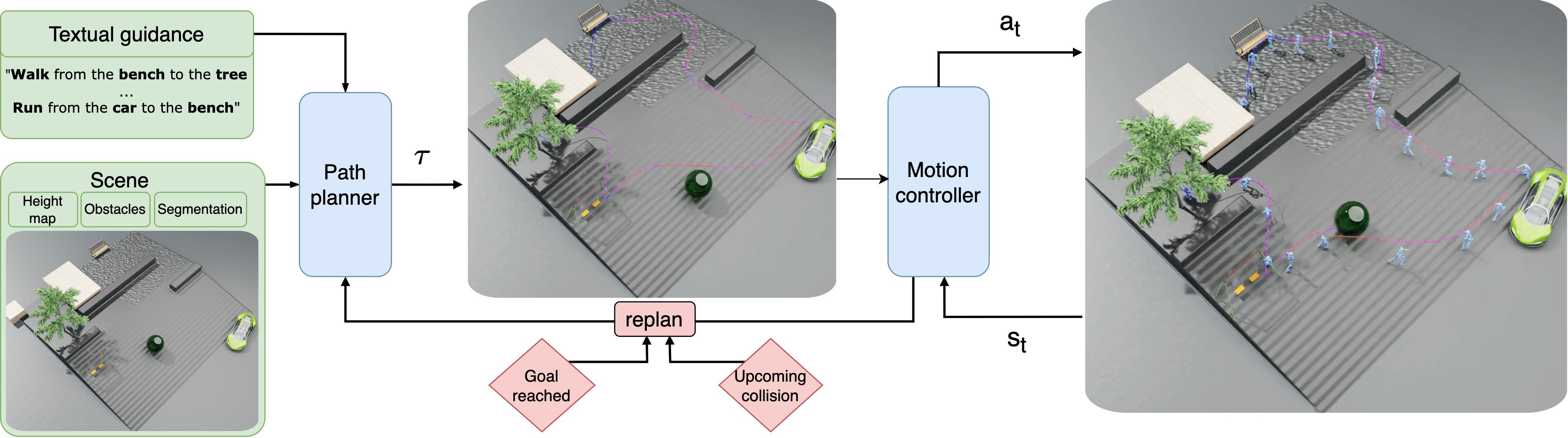}
    \caption{\textbf{Method overview.}  %\gal{Lets use input green boxes consistently. sampels should be both outside (or inside)} 
     A simulated scene is provided to the path planner, together with a series of textual instructions, requesting a humanoid to reach landmarks in the scene using various locomotion types. The high-level planner computes a path that is fed to a reinforcement-learning-based low-level motion controller, which controls a humanoid to follow the path using the request locomotion type.}
    \label{fig:overview}
\end{figure*}

\subsection{3D Scenes}
In this work our characters navigate across rich 3D scenes. We describe the simulated scene as:
\begin{equation}\label{eq:scene}
\textit{Scene}=\{h, \mathcal{O}_\text{static}, \mathcal{O}_\text{dynamic}, \mathcal{O}_\text{top},  \mathcal{L}\} \,.
\end{equation}
The 5 components of the scene are as follows:
\begin{enumerate}
    \item A terrain given as a height map $h(x,y) \in \mathbb{R}$, specifying the height for each location in the grid. 
    \item A set of non-passable static obstacles  $\mathcal{O}_\text{static}$.
    \item A set of dynamic obstacles and their properties $\mathcal{O}_\text{dynamic}$.
    \item A set of height-limiting (``top'') obstacles requiring the agent to crouch or crawl when traversing through them $\mathcal{O}_\text{top}$. 
    \item A set of landmarks $\mathcal{L}=\{L_i\}_{i=0}^l.$    
\end{enumerate}
The locations and speeds of the dynamic objects are updated in each simulation step, based on their provided properties. The goal in these scenes is to plan a path between a series of landmarks $L_1, \ldots, L_k$ and successfully execute the plan.

\section{Method}\label{sec: method}
Our method, \ourmethod, consists of two main components (Figure \ref{fig:overview}):

\textbf{High-level path planner.} The planner receives two inputs: a physically simulated 3D scene and a templated text instruction describing a sequence of (locomotion type, start point, target point). The planner then outputs a path $\tau$ consisting of velocity and 3D head positions at each time-step. The path is optimized to lead the agent through a smooth, physically feasible path to reach the goal while avoiding obstacles. It is then sent to the motion controller.

%\gal{The following are not in figure 2, its explained in 4.2. I would remove this paragraph.} The planner consists of several sub-modules: (a) An A* algorithm on the granulated terrain, (b) a path refiner applying smoothing and obstacle avoidance, and (c) a speed controller for finding the optimal speed along the trajectory. 

\textbf{Low-level locomotion controller.} This controller receives as input the height map $h$ of the local terrain and a desired path $\tau$. It outputs a sequence of motor actions for the humanoid joints, resulting in the humanoid following the desired path. The locomotion controller is trained on real-world motion captures using RL to generate realistic human-like behavior. 

\textbf{Together}, at each time-step of the simulation, the %high-level dynamic 
path planner translates the current humanoid location and target landmark to a local path that the controller aims to follow. The controller then outputs actions on the joints to the simulator, which then advances to the next simulation step. %A diagram of our full system is given in Figure \ref{fig:overview}.

\begin{figure*}[ht]
\includegraphics[width=\textwidth]{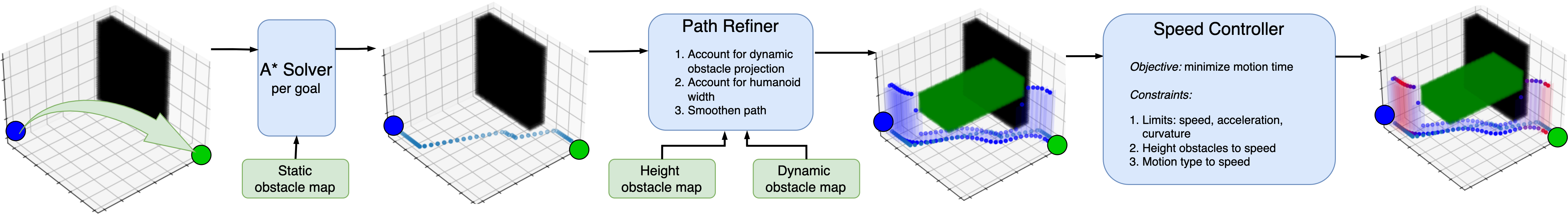}
  \caption{The three stages of our dynamic path planner: (i) $A^*$ solver, (ii) path refiner, and (iii) speed controller.}
  \label{fig:planning_diagram}
\end{figure*}

\subsection{Dynamic Path Planner}
\label{sec:planner}
We present a comprehensive module for path planning, structured into distinct phases. 
The full planning flow is visualized in Figure~\ref{fig:planning_diagram}.% (see also Figure \ref{fig:overview} for context). 

\textbf{Textual guidance parser.}
The input to the planner is a 3D simulated scene $\textit{Scene}$ and 
a series of $k$ text instructions in the form ``Run from the tree to the lake''. 
These instructions refer to known landmarks in the scene where we assign each landmark to its corresponding set of $(x,y)$ coordinates. 
We then translate each instruction to a triplet of (source landmark, target landmark, locomotion type). Together, they form a sequence of instructions that go through the series of landmarks $L_1, \ldots,  L_k$ connected by paths. We observe that two main characteristics that differentiate between desired forms of locomotion are speed and height. As such, each path is represented as a series of 3D positions. We determine velocities by controlling the distance between the points on the path. These paths are fed to the motion controller.

%The output of our planner is a series of $k$ paths.
%We observe that two main characteristics differentiate between desired forms of locomotion are speed and height. As such, each path is represented as a series of 3D positions. We determine velocities by controlling the distance between the points on the path. These paths are fed to the motion controller. 

Our planner is inspired by state-of-the-art robotics and autonomous driving techniques \cite{wang2020robot,hsu2020reinforcement,alcantara2021optimal}. It operates in three stages: low-resolution path construction using the A* algorithm, high-resolution height trajectory creation considering 3D obstacles, and locomotion velocity setting based on path slope and curvature. It ultimately produces a sequence of head positions and velocities for the motion controller, being the first long-horizon planner aware of complex 3D environments and character motion envelopes.

% The planner operates in three steps. \gal{Add here tha we borrow techniques from AV, with refs. Explain why we have three steps}

\textbf{A* with awareness to slope.} We form a 2D grid on the terrain level, connecting every point to its eight neighbors. We then remove connections based on walls, top obstacles too low to bypass, and infeasible slopes. The remaining connections are assigned a weight based on the distance and height difference, representing how difficult and time-consuming it is for the agent to cross it. Then, for every landmark ${L_i}$, we mark all its assigned $(x,y)$ coordinates as goals for A*. Finally, the A* algorithm \cite{bell2009hyperstar} constructs a shortest-path solution to the nearest coordinate of the landmark. To support long paths across large areas, we implement A* efficiently using KD-trees \cite{zhou2008real} and perform it once.

\textbf{Path refiner.} The path created at the previous A* phase is coarse and impervious to dynamic obstacles and height constraints. We refine it to achieve smoother paths that avoid all obstacles. Thus, these paths are easier for the low-level controller to follow, increasing its success rate. 
For \textit{dynamic obstacles}, we detect potential collisions by estimating the movement pattern of the obstacle and comparing it with the last designated path $\tau$.
If a collision is likely to occur within a $1.5 \left[s\right]$ time frame, we replan the path, with the dynamic obstacle set at its future location as an added constraint. Finally, the path refiner adjusts the trajectory for smoothness and obstacle avoidance. 
The path refiner then sets the desired head height (z-axis) values that guide the character beneath top obstacles.

\textbf{Motion-aware speed controller.} 
The path created in the previous phases contains only spatial information and lacks consideration of the humanoid character's performance envelope. The third step assigns a velocity vector to each frame in the path, consistent with the requested motion type (walk, run, crouch, crawl) and the terrain. Specifically, a quadratic program (QP) is solved to minimize travel time along the refined path while avoiding infeasible areas. The QP is constrained by valid speed ranges for each locomotion type and by bounds on longitudinal and lateral accelerations that the character can achieve. The speed controller accounts for path curvature, adapting longitudinal speed based on lateral acceleration and adjusting the speed according to elevation and slope. This results in a refined speed profile that ensures the character slows down before turns and around obstacles.
% \chen{Maybe call this motion awareness?}
% The path created at the previous two phases contains only spatial information and is blind to the performance envelope of the humanoid character. The third step aims to assign a velocity vector to each frame in the path, which is consistent both with the text guidance (different locomotion types have different speeds) and with the terrain (slowing down during steep turns). Specifically, the third phase solves a quadratic programming problem (QP) designed to minimize travel time along the refined path while avoiding impassable areas. The QP is constrained by limits on valid speed range for each locomotion type and by bounds on longitudinal and lateral accelerations that the humanoid character can achieve. This phase accounts for path curvature and adapts longitudinal speed based on the lateral acceleration. Subsequently, the agent slows down before turning. The speed controller also adjusts the speed according to the elevation at each point in the path, resulting in a speed profile for each point on the refined 3D path.

% \gal{This reflects how the planner depens on the controller. It should be in the section about the controller. Not here.  There we should say something like: "The speed assigned along a path is based on the requested motion type (walk, run) the slope (calculated from the height map) and is later refined for slowing down around osbatcles" } \chen{moved this comment to the planner. the "speed controller" looks like a good way to phrase the speed controller part.}

\textbf{Dependence between control and planning:} The effectiveness of our method hinges on the strong interdependence between the control and planning phases. 
% The planner’s output must respect the character’s feasibility envelope, ensuring that all planned movements are within the character's physical capabilities. 
The A* algorithm incorporates height differences into its cost function, considering the difficulty and time required for the character to traverse different elevations.  This notion can be easily extended to different agents with different capabilities; for example, an accomplished humanoid climber will choose a steeper, shorter path than an agent having difficulties with high slopes. The tight coupling continues through the path refinement and speed control phases, where adjustments for slope, curvature, and terrain are directly informed by the character’s performance limits. This ensures that the final path is not only theoretically optimal, but also practically executable, maintaining the balance and stability of the character in real-world conditions.

\begin{figure}[t]
    \centering
    \includegraphics[width=1.0\linewidth]{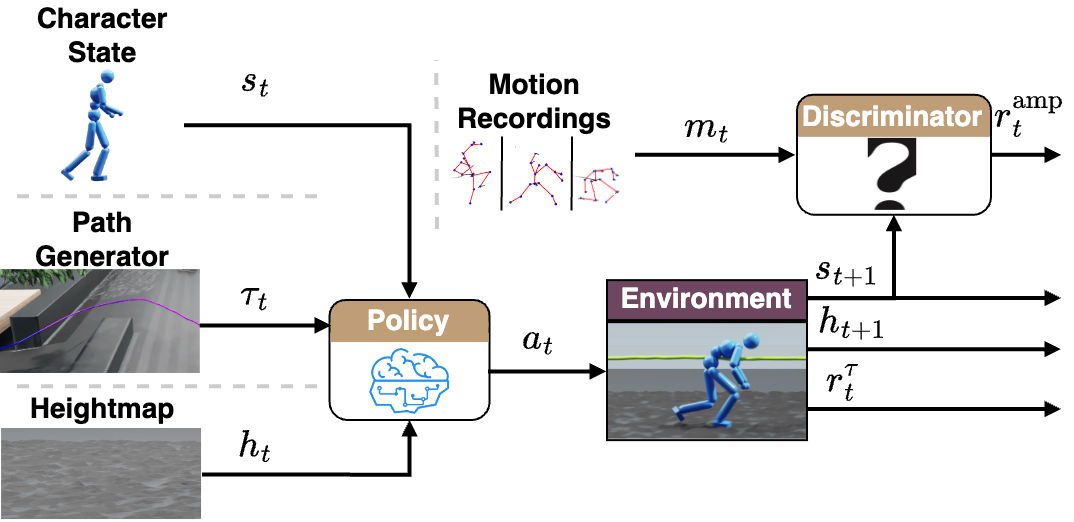}
    \caption{\textbf{Locomotion controller module.} The locomotion policy observes the character state $s_t$, the height map of the terrain $h$, and the requested path $\tau$. During training, the trajectories are randomly sampled. In inference, the trajectories consider the terrain characteristics, such as obstacles. After simulating the predicted action the resulting state is used for providing the style reward $r^\text{amp}$ and for the path following reward $r^\tau$.}% \gal{It would be useful ot make this figure larger. Perhaps make it shorter in height so we can put it full page? or the opposite, make it more narrow, so it takes more text lines? }}
    \label{fig: locomotion}
\end{figure}

\subsection{Locomotion controller}
\label{sec:controller}

Plans produced by the high-level path planner are provided to a low-level locomotion controller. This controller, illustrated in Figure~\ref{fig: locomotion}, is tasked with executing the provided plan using human-like physical motion. The provided plans control the height, location, and speed at which the character should move. This is performed across a wide range of irregular terrains. To achieve this we train our locomotion controller using reinforcement learning. We provide a reward with two components: one for following the path and the other for producing realistic motion. In this paper, we accommodate four forms of locomotion: running, walking, crouch walking, and crawling. It is interesting to note that the planner does not explicitly guide a specific locomotion type, like running or crouching; but since the motion controller is tasked with creating a natural-looking movement, the right type of locomotion naturally emerges for every speed and head height. 

% To support higher-level planning, our work is focused on locomotion in 3D, thus allowing the common point-to-point movements required for the mentioned use cases. We depict the locomotion controller in Figure~\ref{fig: locomotion}. It is constructed to generate human-like physical motion. It needs to do so while adhering to the requested motion traits and traversing diverse nontrivial terrains. For that purpose, we train an RL agent. Our reward function is composed of two main components -- one for following the path and the other for producing realistic motion. We also include two additional rewards, to encourage the agent to look in the direction he is heading, and the other to encourage keeping the head higher than the other body parts. We added the latter since otherwise,  paths looks less realistic. In this paper, we accommodate four forms of locomotion: running, walking, crouch-walking, and crawling. 

\textbf{State.} At each step the controller observes the current pose of the character $s_t$, the target trajectory $\tau_t$, and the surrounding heightmap $h_t$. The trajectory is provided as 5 seconds into the future with ten samples, one every 0.5 seconds. The height map allows the controller to be aware of its surroundings, ensuring motions are robust to varying scenes. We represent the height map as a rectangular grid rotated to the character's root orientation. It consists of 64 equidistant samples spanning $0.8 \left[m\right]$ on each axis.

\textbf{Reward.}  The goal of the controller is to track a target path $\tau$, generated by the planner (Section~\ref{sec:planner}). We identify that speed and height are the two main characteristics for controlling our desired forms of locomotion (run, walk, crouch-walk, crawl). As such, the path is constructed in 3D. At each time step $t$, the point $\tau_t$ represents the 3D coordinates for the head. By controlling the distance between the points, we determine the requested speed, and changing the $Z$-axis provides control over the desired character's height.

Due to the importance of height control, we split the reward into four components. The first considers the $x$-$y$ displacement of the character. This guides the path the character should follow, in addition to speed. The second component considers the height of the head. The third considers the head's direction (pitch and yaw), aligning it with the path. Since all path rewards target the head, to produce more realistic movement when following lower heights we add the reward term: $r^{\text{body}}_t=c_\text{body}(z_\text{head} - \max_\text{body}z)$ with $c_\text{body}=0.1.$ This term encourages the head to be the highest point to prevent a hunchback posture when crouching and crawling.
\begin{equation*}
    r_t^\tau = e^{- c_\text{pos} || \tau_t^{x,y} - s_t^{x,y} ||^2} + e^{- c_\text{height} || \tau_t^z - s_t^z ||^2} + e^{-c_\text{dir}||\tau_t^{y,p} - s_t^{y,p}||^2} + r^{\text{body}}_t\, ,
\end{equation*}
where $c_\text{pos}=2$, $c_\text{height}=10$ and $c_\text{dir}=20$ are fixed. 

Complementing the task reward, we leverage AMP \cite{peng2021amp} for motion stylization. AMP trains a discriminator to differentiate between reference recorded motions $m_t$ and those simulated by the motion controller $s_t$. The resulting reward combines the tracking objective with the discriminative reward: $r_t = r_t^\text{amp} + r_t^\tau \,.$

\textbf{Reference motions.} The adversarial reward in AMP is maximized when the simulated pose distribution matches the data. Naively providing a diverse dataset of human motions may result in unwanted behaviors. For example, back-flips and break-dancing are distinctly different from locomotion. To ensure that the motion distribution matches the motions needed to solve the task, we hand-picked 75 distinct motions from the AMASS dataset \cite{AMASS:ICCV:2019}, covering the four locomotion forms performed in various styles. An appropriate ablation study is provided in Section \ref{sec: experiments}.

% The correlation between data and task works both ways. 
\textbf{Random path generation.} During training, the path generator samples random paths for the character to follow. If the task demands unlikely motion given the data, we observe a degradation in both motion quality and tracking performance. For example, crawling at $v=5 m/s$ is unrealistically fast and will not appear in our motion capture data. Thus, the task and discriminative rewards will contradict each other, impairing the training process. To provide a realistic path, the generator first samples the requested heights along the path, ranging from $0.4\left[m\right]$ (crawling) to $1.47\left[m\right]$ (the simulated humanoid's height), with smooth transitions in between. Then, the speed is sampled uniformly between $0$ and $v_t^{\max{}}$, where:
\begin{equation*}
    v_t^\text{max} = \min{(1 + 4 \cdot (z_t - z^\text{min}) / (1.2 - z^\text{min}),\, V_{\max{}})} \,,
\end{equation*}
for a minimal height $z^\text{min} = 0.4\left[m\right]$ and maximal speed $V_{\max{}}=5 \left[m/s\right]$. This reasults in a linearly increasing max-speed, ranging from $v_t^\text{max}=1 \left[m/s\right]$ for crawling motions, and up to $v_t^\text{max}=5\left[m/s\right]$ for any height above $1.2\left[m\right]$.

\textbf{Actions.} Similar to prior work \cite{peng2018deepmimic,rempe2023trace}, we opt for proportional derivative (PD) control. We represent the action distribution using a multi-dimensional Gaussian with a fixed diagonal covariance matrix.

% To encourage the correlation of reward components \gal{?}, we ensure a tight coupling between the requested height and speed. First, the path generator samples the requested height for each point along the path. 

% To provide a realistic path, the generator first samples the requested heights along the path, ranging from $0.4\left[m\right]$ (crouching) to $1.47\left[m\right]$ (our humanoid's height is $1.47 \left[m\right]$), with smooth transitions in-between. Then, the speed is sampled uniformly between $0$ and $v_t^{\max{}}$, where:
% \begin{equation*}
%     v_t^\text{max} = \min{(1 + 4 \cdot (z_t - z^\text{min}) / (1.2 - z^\text{min}),\, V_{\max{}})} \,,
% \end{equation*}
% for a minimal height $z^\text{min} = 0.4\left[m\right]$ and maximal speed $V_{\max{}}=5 \left[m/s\right]$. This reasults in a linearly increasing max-speed, ranging from $v_t^\text{max}=1 \left[m/s\right]$ for crawling motions, and up to $v_t^\text{max}=5\left[m/s\right]$ for any height above $1.2\left[m\right]$.

\section{Experiments}\label{sec: experiments}

% Explain what we will show in the experiments

% Complementing the numerical evaluations, we provide video demonstrations and comparisons in the supplementary material.

% Explain about the data

% Explain about the humanoid model. Can copy from CALM/PADL/ASE.
We test \ourmethod{} both qualitatively and quantitatively. 
First, we evaluate the motion controller with several common forms of motion across a range of randomized terrains. In addition, we evaluate our design decision regarding the workflow for training the controller, using ablation experiments regarding both data curation and height-speed coupled path generation.

Then, we analyze the quality of the path planner. We focus on evaluating the paths produced by the QP solver in terms of how well they fit the capability envelope of the locomotion controller.
Finally, we provide qualitative examples both for the locomotion controller and the whole system in a complex scene with multiple landmarks. We stress that our test scenes were not observed during training. Our agent can navigate during inference in every scene as defined in \eqref{eq:scene}, without requiring any further training.
We provide video demonstrations and comparisons in the supplementary material. 

\subsection{Experiment details}
We focus here on the SMPL humanoid \cite{SMPL:2015} with a neutral body structure. The humanoid is simulated in IsaacGym \cite{makoviychuk2021isaac}, operating at $120 \left[Hz\right]$. The controller is trained using PPO \cite{schulman2017proximal} and takes decisions at a rate of $30 \left[Hz\right]$. The motion controller is trained for one week on a single NVIDIA V100 GPU. 

% \textbf{Simulation.} To simulate our physical environment, we used IsaacSim. %\footnote{NVIDIA Isaac Sim. Available online: \hyperlink{https://developer.nvidia.com/isaac-sim}{https://developer.nvidia.com/isaac-sim}}. 
% The controllers make decisions at a rate of $30 \left[Hz\right]$, while the simulation runs at $120 \left[Hz\right]$. We train the motion controller for 7 days on a single NVIDIA V100 GPU. 

\textbf{Training Data.} We utilize motion capture data from the AMASS dataset \cite{AMASS:ICCV:2019} that contains more than 10 hours of diverse motion recordings across thousands of distinct motions. Complementing AMASS, HumanML3D \cite{guo2022generating} provides textual labels for each motion.

\section{Results}
We now discuss the resulting paths and motions produced by \ourmethod{}. 

\subsection{Locomotion}\label{exp: locomotion}

%\gal{When cutting down, keep results, shrink excessive text. Can we add figures showing planned paths?}

We test our locomotion module on various terrains for every motion type; \textbf{example videos are given in the supplementary material}, and screen captures are shown in \cref{fig: locomotion example}.  \cref{tab: locomotion controller}, reports the XY-displacement along the path, which shows how well the controller tracks the requested path, and the Z-displacement, which measures how well it maintains the requested height. The results show a hierarchy of complexity between tasks such that stairs are harder than rough gravel-like terrain, which is harder than locomotion across a flat and smooth sloped surface. Although the data only contained motions traversing flat, smooth floors, the controller learned to generalize to a wide range of uneven terrains and produce robust motions resembling those in the data. Also, as expected, moving at lower speeds enables the character to follow the path more precisely, thus maintaining a lower error. 
% \gal{I suggest to remove the remaining of this par'} \galst{Notice that crouch-walking has a very small Z error. We attribute this to the phase change between maintaining a very low height while standing and reducing further in height, which requires kneeling. Hence, crouch-walk is a highly constrained motion yielding low variability in the head's height. }

\paragraph{Locomotion ablation} We consider the crawling on rough terrain. We focus on this task as it was shown in \cref{tab: locomotion controller} as the hardest type of motion to generate. For our ablation study, we consider two design decisions -- data curation, and random path generation. \textbf{Data curation:}\footnote{We provide additional information in the supplementary material.} We evaluate the importance of data curation by comparing training on the full dataset (\textit{none}), automated curation (\textit{naive}, based on the HumanML3D \cite{guo2022generating} textual labels), and manual curation (\textit{curated}). \textbf{Random path generation:} To showcase the importance of coupling height and speed in the randomly sampled paths, we also compare our \textit{coupled} motion-aware path sampler that samples the requested speed along a random path according to the requested height, with a \textit{naive} baseline of uncoupled sampling of both speed and height.

The results, shown in \cref{tab: locomotion ablation}, emphasize the importance of these design decisions. We find that the best results are obtained by combining the manually curated dataset, alongside the coupled random path generator. By ensuring a tight alignment between the requested motions, the data representing them, and the randomly generated paths, we find an improvement of 50\% in the XY-err, compared to the next best combination.
% although training on the entire AMASS dataset is possible, many motions in that dataset are irrelevant to locomotion, such as performing hand gestures or dancing.  Since the reward in our model is discriminative in nature \cite{peng2021amp}, training with a data distribution that does not match the task distribution may result in uncanny or even bad motions. In this work, we tested two methods of data purification, which we compare here in an ablation study. 

\begin{table}[t]
    \centering
    \caption{\textbf{Locomotion ablation.} We compare various design choices for data purification and path generation during training. The analysis focuses on crawling on rough terrain, with additional results provided in the supplementary material. The results were averaged over $10^6$ runs.}
    \begin{tabular}{l|c||c|c}
                    % & \multicolumn{2}{c}{\textbf{Crawl}}  \\ \toprule 
                   Data purification & Path generator & XY-err & Z-err \\ \toprule
         \multicolumn{2}{c|}{\textbf{Ours} (Curated + Coupled)} & 0.25 & 0.10 \\ \hline
         Curated & Naive &  0.50 & 0.13 \\ \hline
         Naive & Naive  & 0.78 & 0.09 \\ \hline
         Naive & Coupled  & 1.16 & 0.17 \\ \hline
         None & Coupled  & 0.66 & 0.09 \\ \hline
         None & Naive  & 1.39 & 0.19
    \end{tabular}
    \label{tab: locomotion ablation}
\end{table}

\begin{table*}[ht]
    \centering
    \caption{\textbf{Mean path displacement for various locomotion types .} We test four motion types over four types of terrains. For each, we sample a path with constant speed and height, and compute the mean error between the position of the humanoid and the desired path. The results were averaged over $10^6$ runs.}
    \begin{tabular}{l||c|c|c|c|c|c|c|c}
                    & \multicolumn{2}{c|}{\textbf{Crawl }} & \multicolumn{2}{c|}{\textbf{Crouch-walk}} & \multicolumn{2}{c|}{\textbf{Walk}} & \multicolumn{2}{c}{\textbf{Run}}  \\ \toprule 
                   \textbf{Terrain} & XY-err $\left[m\right]$ & Z-err$\left[m\right]$ & XY-err $\left[m\right]$& Z-err $\left[m\right]$& XY-err$\left[m\right]$ & Z-err $\left[m\right]$& XY-err$\left[m\right]$ & Z-err$\left[m\right]$ \\ \hline
        Flat    & 0.20 & 0.07 & 0.20 & 0.03 & 0.24 & 0.21 & 0.52 & 0.26 \\ \hline
        Slope   & 0.21 & 0.09 & 0.22 & 0.03 & 0.25 & 0.21 & 0.60 & 0.27 \\ \hline
        Rough   & 0.25 & 0.10 & 0.25 & 0.04 & 0.29 & 0.24 & 0.79 & 0.31 \\ \hline
        Stairs  & 0.45 & 0.13 & 0.41 & 0.07 & 0.39 & 0.27 & 1.01 & 0.35
    \end{tabular}
    \label{tab: locomotion controller}
\end{table*}

\subsection{Terrain-dependent path planning}
\begin{figure}[ht]
    \centering
    \includegraphics[width=1.0\linewidth]{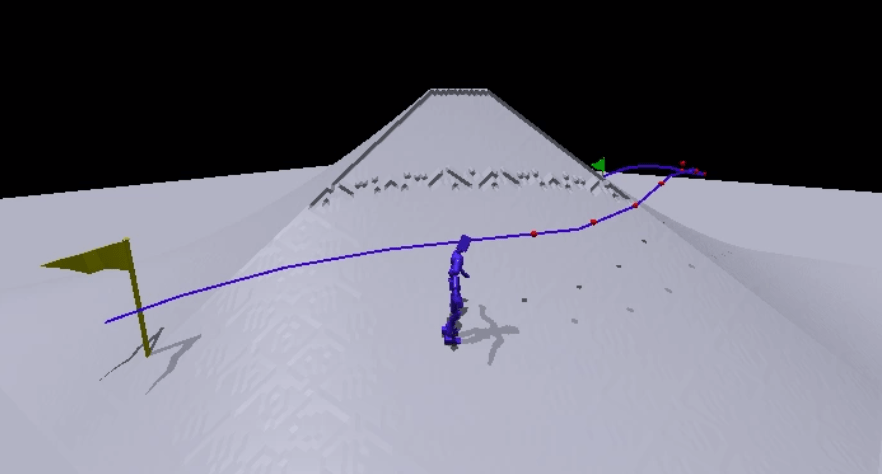}
    \caption{\textbf{Terrain dependent path planning} The humanoid does not go through the shortest path between the two landmarks but prefers a smoother, flatter surface due to the height differences. }
    \label{fig: pyramid}
\end{figure}

When animating a character we often want it to exhibit behaviors that align with the character's story. For example, while a young athlete may be capable of running up a hill, an old person may prefer to take a longer and less physically challenging path. Here, we show that these preferences can be encoded into our path planner. To do so, we re-weight the connectivity graph based on the slope. The distance between each two points is increased by $e^{c \cdot \text{slope}} - 1$. As illustrated in \cref{fig: pyramid}, as a result of increasing the weight, our planner produces a path around the hill instead of directly across it.

%Difficult terrain can cause the humanoid to move slower, failing to follow the requested 3D path, or worse yet -- to crash and terminate. To reduce such failures we embed the difficulty into the planning by changing the weights of the A* algorithm. The distance between two neighboring points with a large slope between them is artificially increased by $e^{c \cdot \text{slope}} - 1$ where $c$ is tuned according to the low-level motion controller capabilities. In Figure~\ref{fig: pyramid}, we show how this added weight encourages the humanoid to take a detour around the hill to reduce time and risk.

\subsection{Speed control}
The third phase of our path planner uses a QP solver to optimize the speed along the path curves. On the one hand, if the speed is too high for the curvature, the character will not be able to maintain the desired route, resulting in large displacement errors and possible collisions with obstacles. On the other hand, if the speed is too low, it takes the character an unnecessarily long time to complete the route. To avoid jerks, the optimization program also considers the maximum acceleration ($a_{\max{}}=0.5 \left[m/s^2\right]$) and deceleration ($a_{\min{}}=-0.1 \left[m/s^2\right]$). 

We demonstrate this trade-off in the ``Slalom'' scene (Figure~\ref{fig: slalom}). This scene requires the character to take steep turns. Here, we compare constant speed versus adaptive speed. We generate both routes using our $A^*$ solver and path refiner. However, we only refine the ``adaptive speed" run using our speed controller module. Without adapting the acceleration profile to the curvature of the path, the humanoid fails to follow the path accurately and often collides with the walls. 

\begin{figure*}\centering
     \begin{subfigure}[b]{0.49\textwidth}
         \centering
             \includegraphics[width=0.87\linewidth]{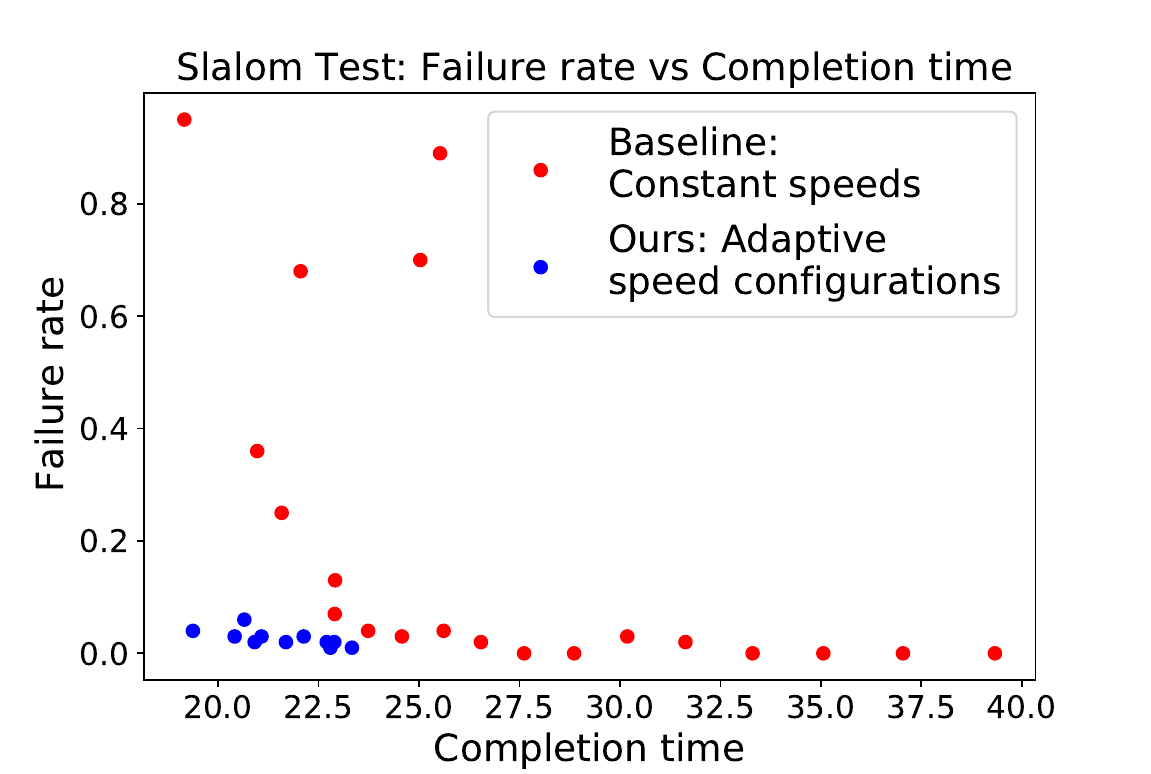}
    \caption{\textbf{Failure rate and Completion Time: Pareto curve in the Slalom test}. Testing our speed controller vs. constant speed. Our planner obtains lower failure rate while retaining faster completion.}
    \label{fig: slalom pareto 1}             
     \end{subfigure}\hfill
     \begin{subfigure}[b]{0.49\textwidth}
         \centering
\includegraphics[width=0.87\linewidth]{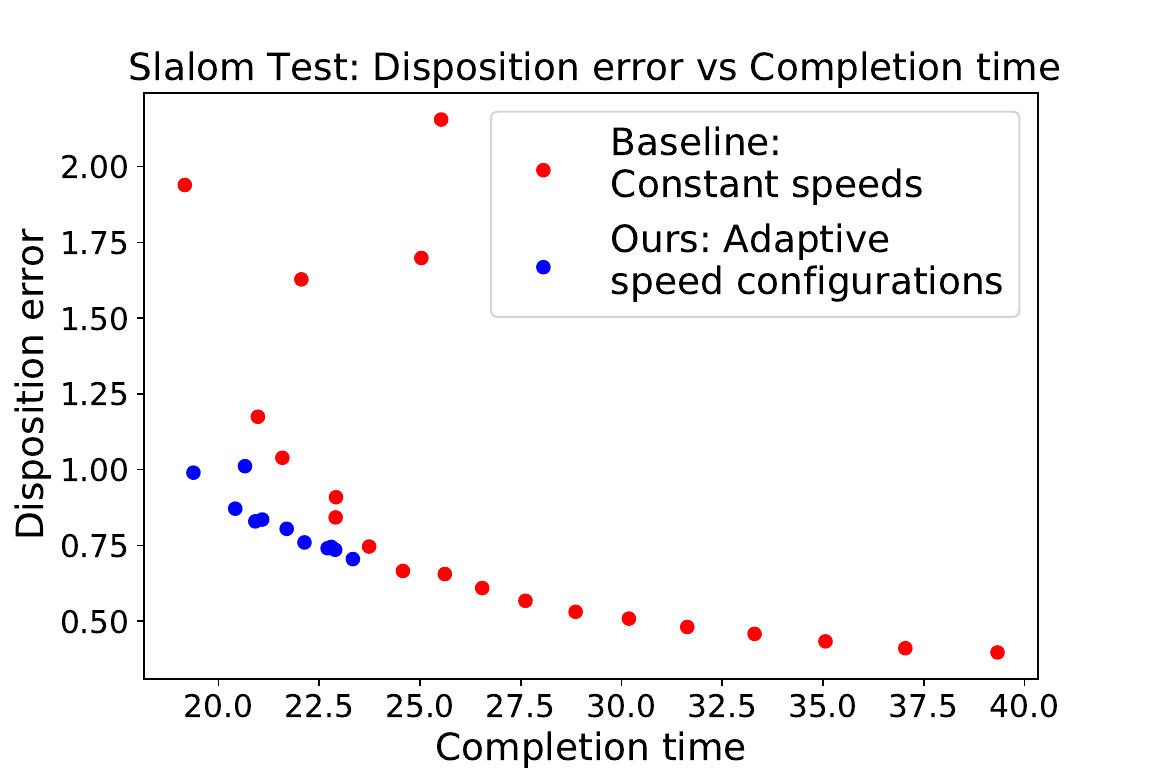}
    \caption{\textbf{Disposition error and Completion Time: Pareto curve in the Slalom test.} Testing our our speed controller vs. constant speed. Our planner obtains lower disposition error while retaining faster completion.}
    \label{fig: slalom pareto 2}
     \end{subfigure}
\end{figure*}

For a quantitative evaluation of the Slalom test, we compare various constant speeds, as well as various configurations of our QP solver. The latter includes different combinations of maximal lateral and longitudinal acceleration. We summarize the results in Figures~\ref{fig: slalom pareto 1} and \ref{fig: slalom pareto 2}. Each choice of constant speed and QP configuration corresponds to a single point in the plots and is the average of 100 runs. The constant speeds were chosen between $1 \left[m/s\right]$ and $3.5 \left[ m/s\right].$ In Figure~\ref{fig: slalom pareto 1}, we report the success rate versus the average completion time. Success is defined as successfully reaching the target landmark. In Figure~\ref{fig: slalom pareto 2}, we show the same, but with the average 3D disposition error on the y-axis. As seen, for every QP configuration we tested, our planner enables the humanoid to traverse its path faster and without compromising on accuracy thus achieving the Pareto front.
% \begin{figure}[ht]
%     \centering
%     \includegraphics[width=0.87\linewidth]{figures/planner/slalom_pareto1.pdf}
%     \caption{\textbf{Failure rate and Completion Time: Pareto curve in the Slalom test}. Testing our speed controller vs. constant speed. Our planner obtains lower failure rate while retaining faster completion.}
%     \label{fig: slalom pareto 1}
% \end{figure}

\subsection{Randomized routes in a complex randomized scenes}\label{exp: full system}
\begin{figure}[ht]
    \centering
    \includegraphics[width=1.0\linewidth]{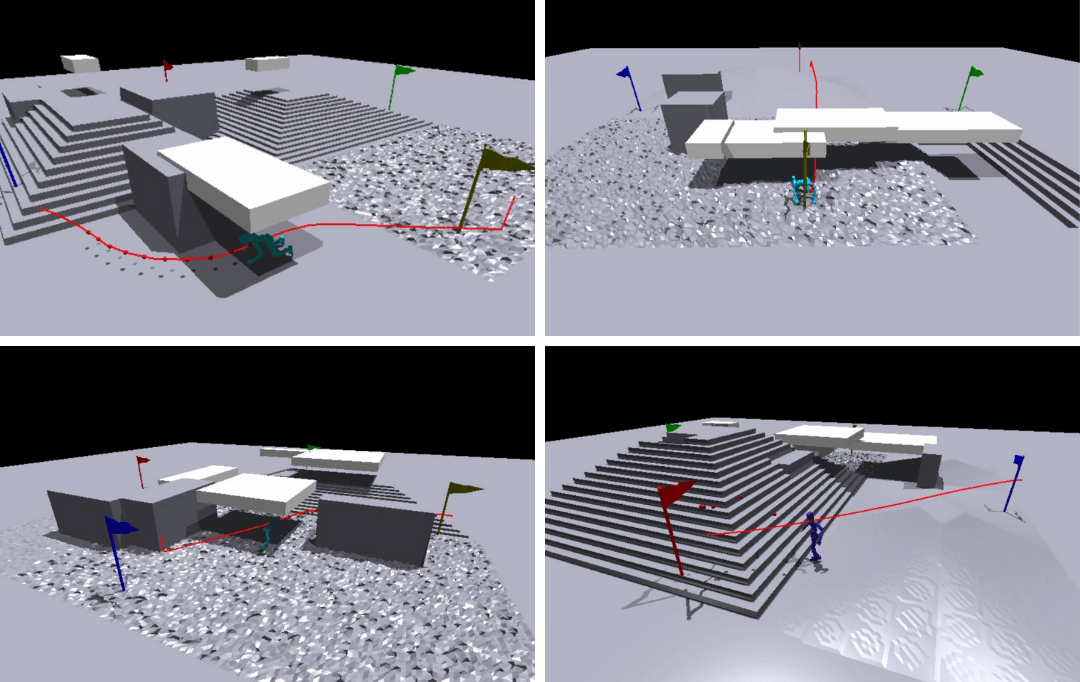}
    \caption{\textbf{Randomized complex scenes} We randomize several terrain types, walls and top obstacles to generate a variety of scenes.}
    \label{fig: randomized}
\end{figure}

%\gal{I revised this}
Finally, we showcase \ourmethod{'s} ability to solve previously unseen scenes in varying forms (Figure~\ref{fig: randomized}). %To do so, w
We automatically construct random scenes consisting of: \textbf{(1)} multiple forms of terrains (smooth, gravel-like, stairs and sloped), with \textbf{(2)} several types of obstacles (walls and low-ceiling obstacles), and four landmarks in each scene. \ourmethod{} is given a random scene with navigation instructions to a random sequence of landmarks. Figure~\ref{fig: randomized} illustrates some of these environments, showing that \ourmethod{} successfully plans realistic routes. In our experiments, our agent has an 81\% chance to fully solve a randomized 4-landmarks scene. % \gal{Do we have any quantitative evidence that it succeeds? }. 
%Examples of several scenes are given in Figure~\ref{fig: randomized}. 

\begin{figure*}
     \centering
     \begin{subfigure}[b]{0.49\textwidth}
         \centering
         \includegraphics[trim={15cm 8cm 12cm 1cm},clip,width=0.9\textwidth]{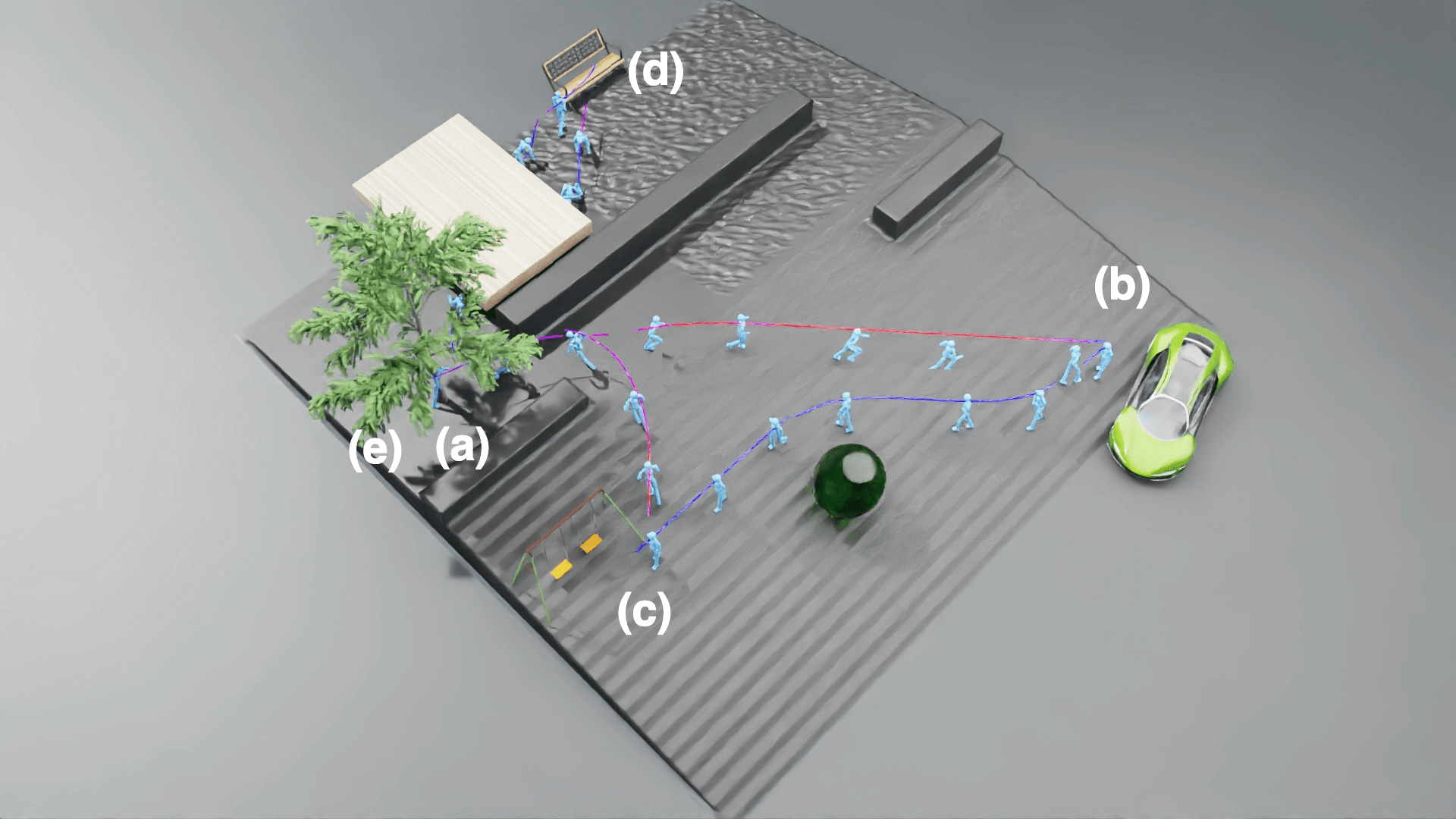}
         \caption{Starting at the tree: "Run to the car" then "walk to the swing" then "run to the bench" then "walk to the tree".}
         \label{fig: rand 2}
     \end{subfigure}\hfill
     \begin{subfigure}[b]{0.49\textwidth}
         \centering
         \includegraphics[trim={15cm 8cm 12cm 1cm},clip,width=0.9\textwidth]{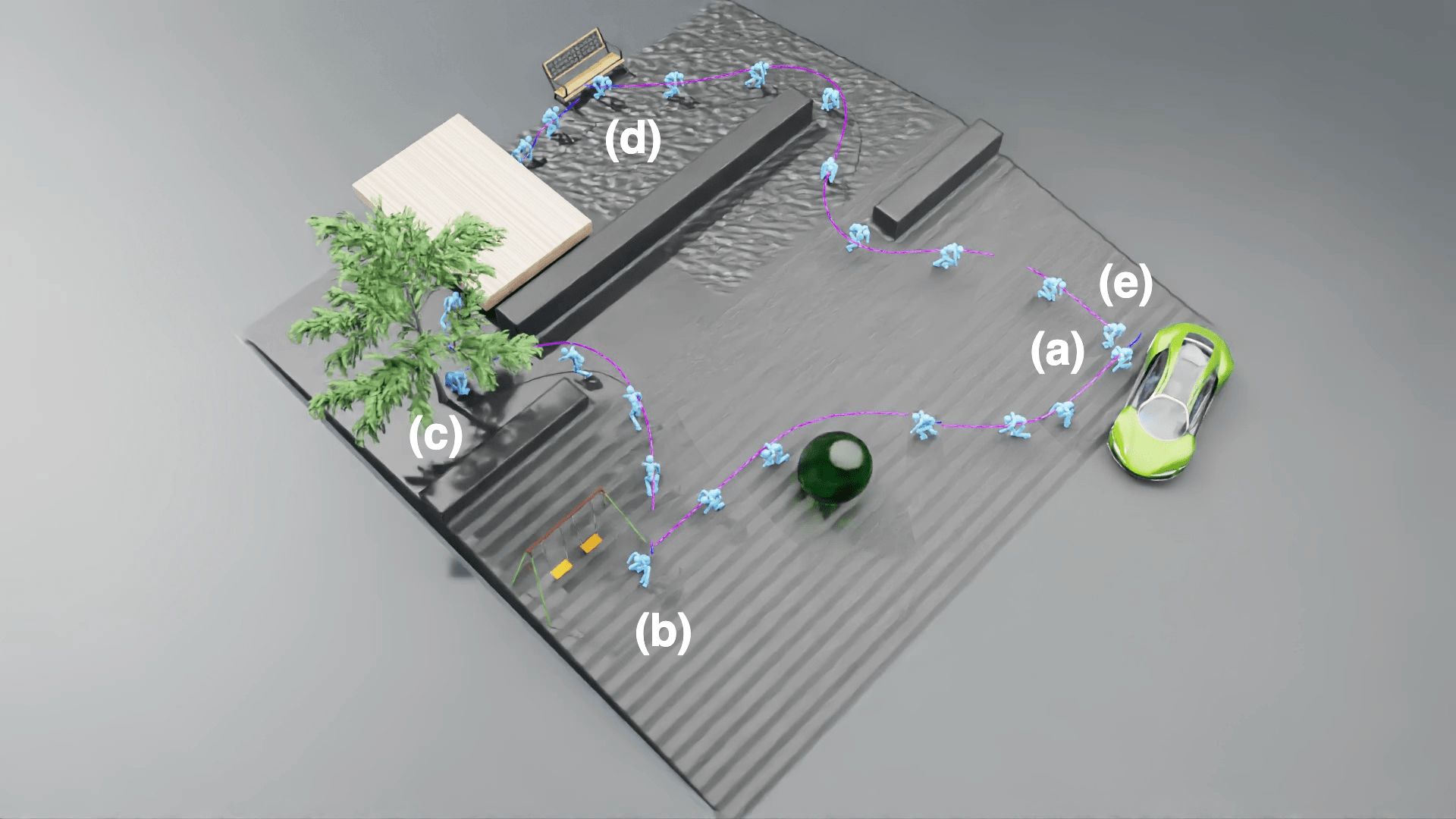}
         \caption{Starting at the car: "Walk crouching to the bench" then "walk crouching to the tree" then "walk crouching to the bench" then "walk crouching to the car".}
         \label{fig: rand 8}
     \end{subfigure}\\
     \begin{subfigure}[b]{0.49\textwidth}
         \centering
         \includegraphics[trim={15cm 8cm 12cm 1cm},clip,width=0.9\textwidth]{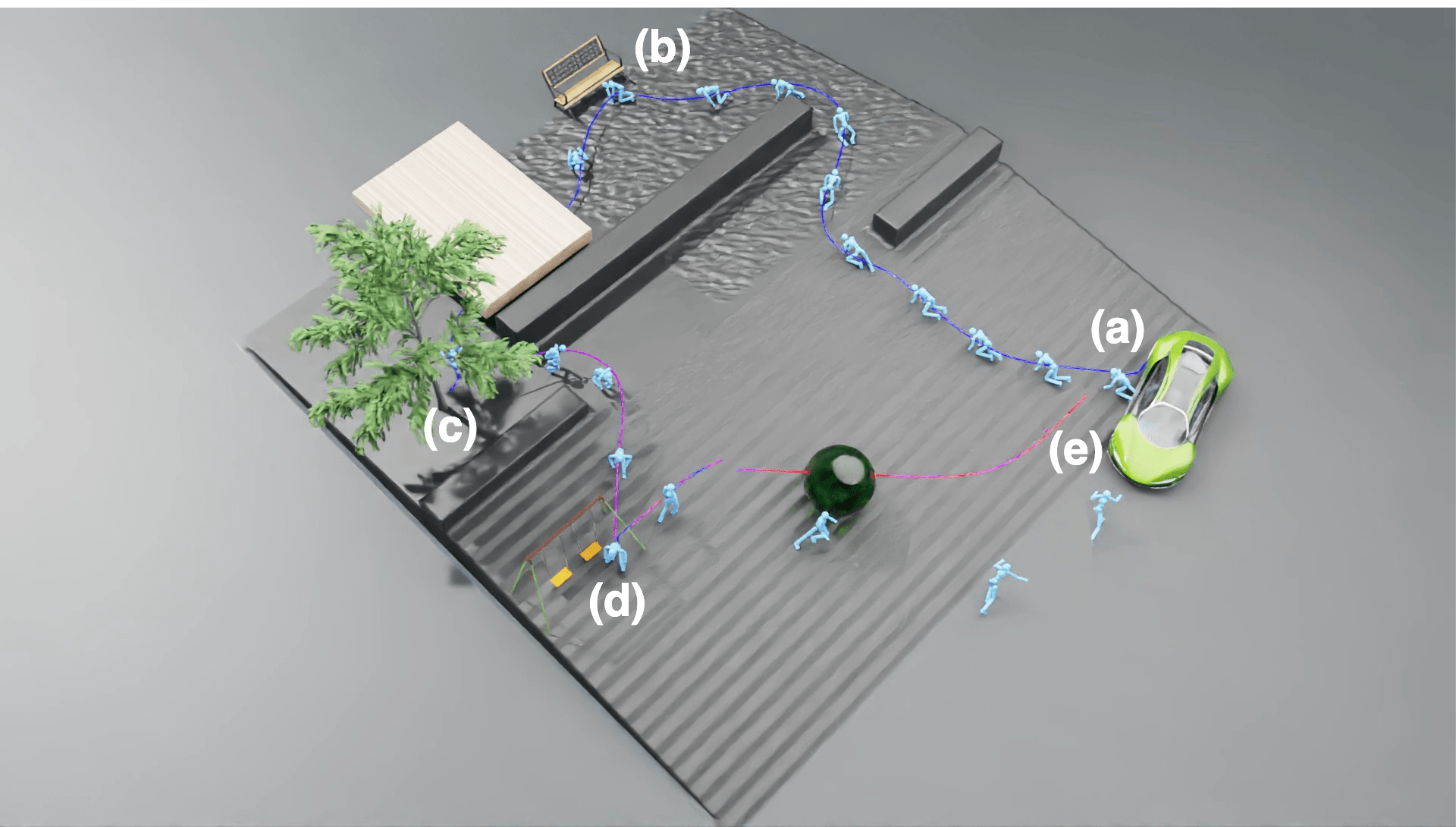}
         \caption{Starting at the car: "Crawl to the bench" then "crawl to the tree" then "walk crouching to the bench" then "sprint to the car"}
         \label{fig: rand 15}
     \end{subfigure} \hfill
     \begin{subfigure}[b]{0.49\textwidth}
         \centering
         \includegraphics[trim={15cm 8cm 12cm 1cm},clip,width=0.9\textwidth]{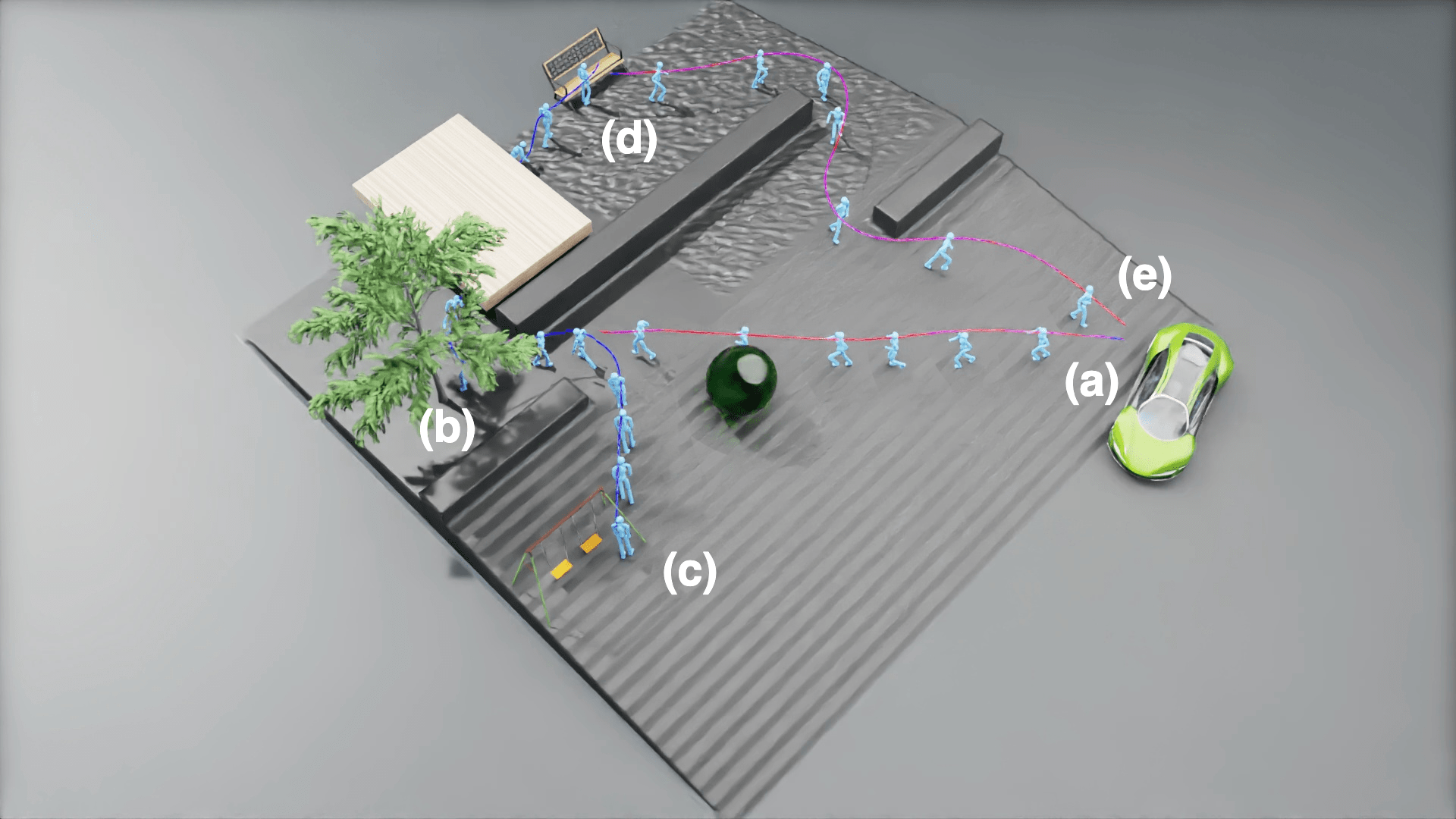}
         \caption{Starting at the car: "Run to the tree" then "walk to the swing" then "walk to the bench" then "run to the car".}
         \label{fig: rand 21}
     \end{subfigure}
     \caption{\textbf{Diverse solutions:} Our system can solve unseen tasks in diverse forms. Here, we show how the character traverses a random sequence of segments within the scene. Each example showcases a randomly sampled sequence of waypoints to be performed with a randomly sampled motion.}% The requested solution, for each segment, is described in the corresponding caption for each figure.}
    \label{fig: playground}
\end{figure*}

As shown in \cref{fig:teaser,fig: playground}, \ourmethod{} solves this task in varying forms while adhering to the scene constraints. It slows down and starts to crawl before passing under the low ceiling (top left), speeds up during straight lines while carefully slowing during a tight curve, and finally, it dynamically adapts the path when a collision with a dynamically moving object is deemed likely\footnote{The dynamic re-planning is best seen in the supplementary video.}.

% To showcase the full scope of our system, we devised a scene that requires the character to succeed in multiple aspects: 
% \begin{enumerate}
%     \item Navigating to various landmarks on the map forming a long scenario without failing.
%     \item Performing diverse types of motions along the route.
%     \item Completing the route quickly while handling sharp turns.
%     \item Avoiding static, top and dynamic obstacles.
% \end{enumerate}
% Our scene and a sample route and choice of motion types are given in Figure \ref{fig:teaser}. We include in the supplementary material multiple videos of the character following randomized routes and motion types in this scene, demonstrating the robustness and potential of our system. Screen captures are given in Figure \ref{fig: playground}.

% Now we add in the path planner. Explain the 3-4 domains we evaluate on.
% Quantitative experiments need to show success rates in solving the task (reaching the end).
% Qualitative experiments need to show (1) how the planner generates meaningful paths that aren't "trivial". (2) how it adapts the height to the 3D terrain. (3) how the character crawls under an obstacle. (4) the system adapts the entire motion to a dynamic obstacle (if we still intend to support this).

\section{Discussion, limitations and future work}\label{sec: limitations}
We described \ourmethod{}, a framework for planning long paths and controlling the motion of humanoid characters in simulated environments for physics-based animation. Given a complex simulated scene and textual instructions, the system efficiently plans a path that a humanoid character can traverse using natural locomotion. The planner takes into account the character's performance envelope in uneven terrain. In addition, it takes into account both static and dynamic obstacles. Then, the trained motion controller successfully tracks these paths, producing complex, scene-aware motions, for example, crawling under a low-hanging ceiling or quickly avoiding an incoming projectile. 

\ourmethod{} can handle various scene complexities, but has some limitations which we outline as future work. \textbf{Interaction with objects and multiple agents. }  Many interesting scenarios for gaming and simulation would require physical interactions with dynamic objects or other agents. Learning such interactions not only requires specific data but may require extending the human model and potentially the controller training, for example with hands \cite{SMPL-X:2019}.  \textbf{Dynamic objects. } \ourmethod{} handles objects whose motion can be easily predicted using rules. Modeling more complex dynamic objects may require learning-based approaches to model the dynamic scene. \textbf{Rich language and scene understanding. } \ourmethod{} focuses on planning and control. It opens opportunities for combining it with modern language models and 3D scene understanding. With these expansions in mind, we see \ourmethod{} as a stepping stone in a much greater system where non-playable characters (NPCs) are given roles to play, forming a rich simulated virtual world.

\newpage

% Bibliography
\bibliographystyle{ACM-Reference-Format}
\bibliography{bibliography}

\newpage

\begin{figure*}
     \centering
     \begin{subfigure}[b]{0.33\textwidth}
         \centering
         \scalebox{-1}[1]{\includegraphics[width=0.33\textwidth]{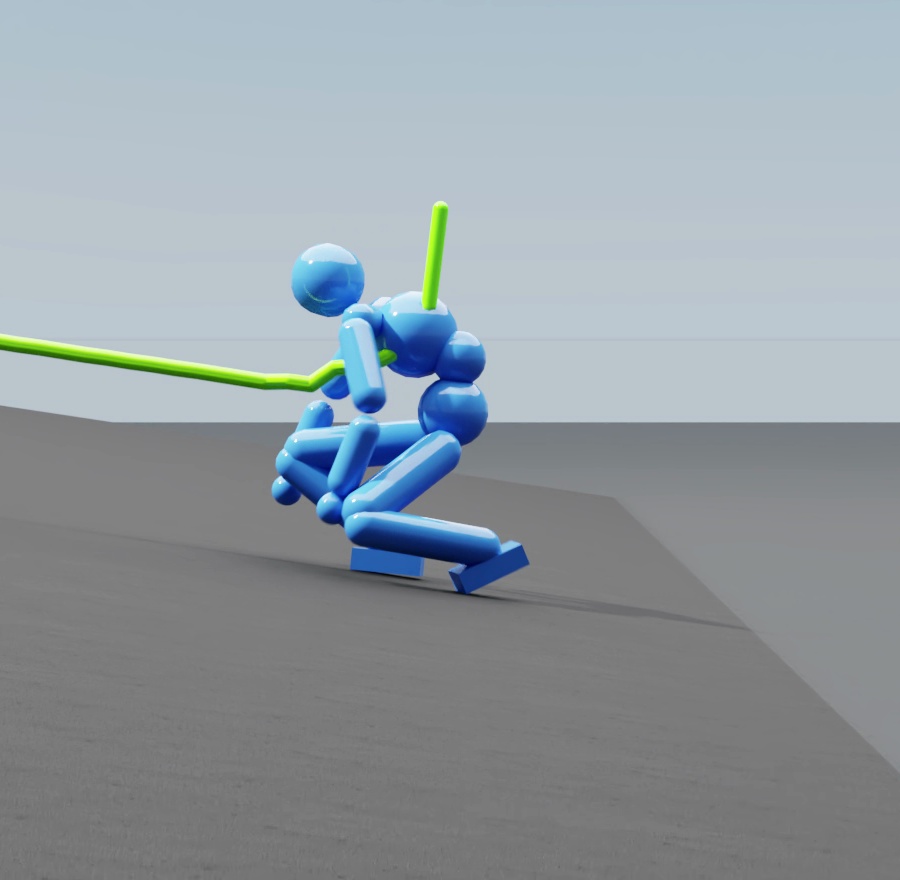}}\hfill
         \scalebox{-1}[1]{\includegraphics[width=0.33\textwidth]{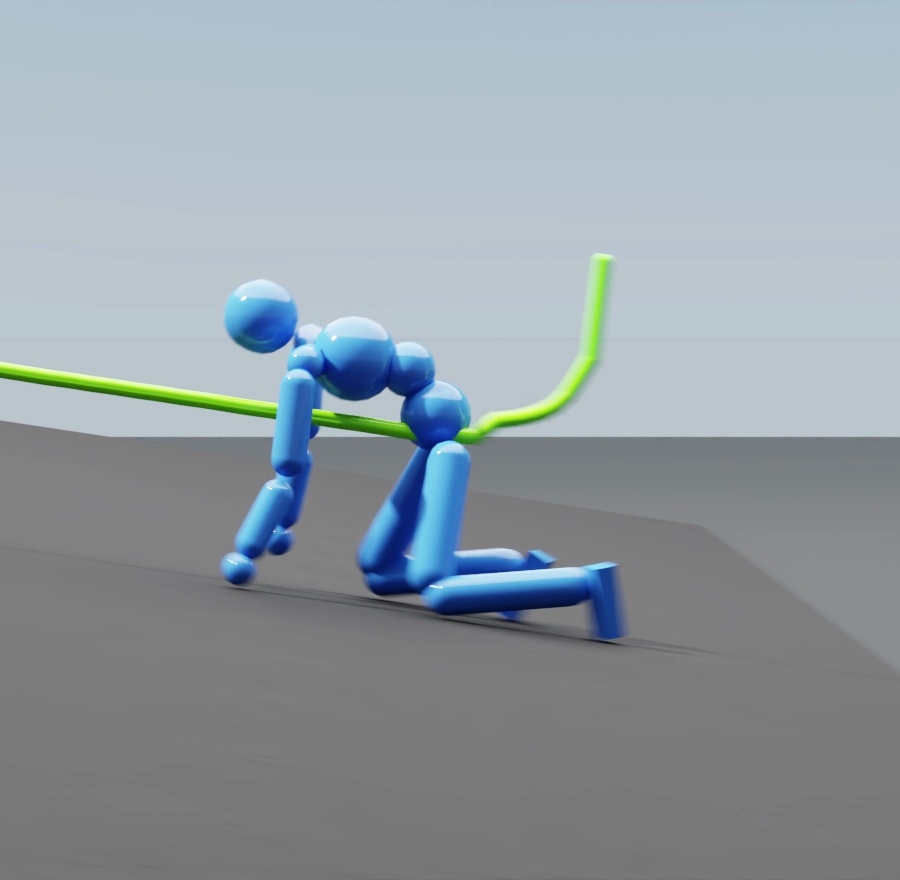}}\hfill
         \scalebox{-1}[1]{\includegraphics[width=0.33\textwidth]{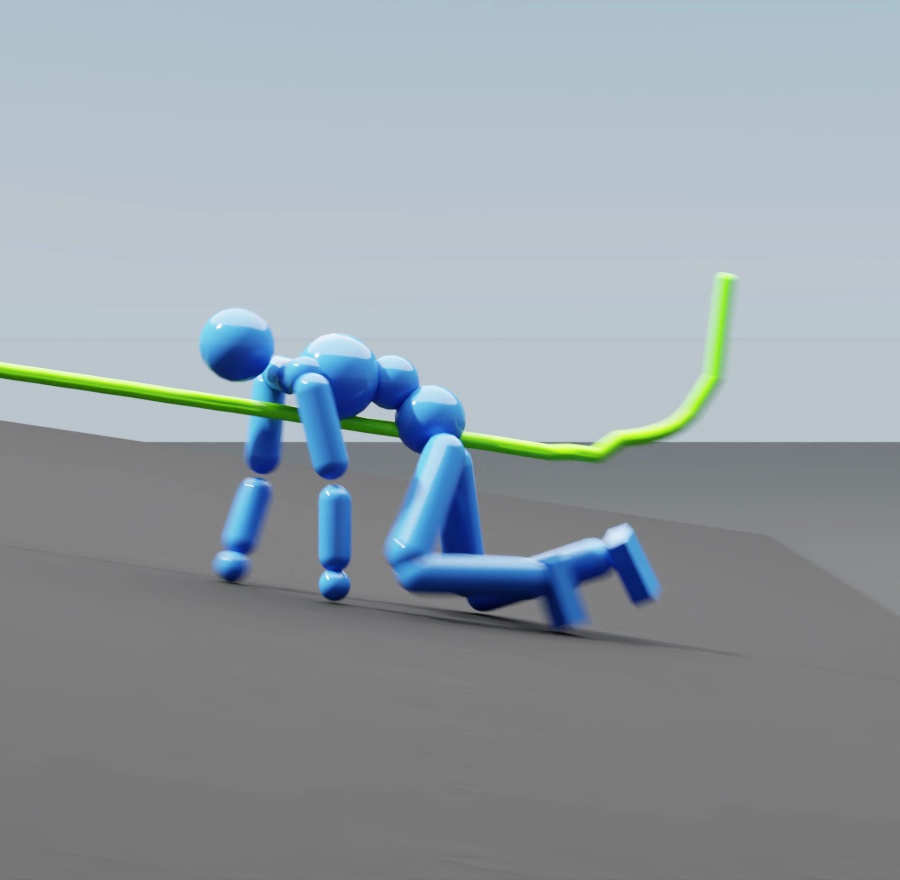}}
         \caption{Crawling up a slope}
         \label{fig: locomotion crawl slope}
     \end{subfigure}\hfill
     \begin{subfigure}[b]{0.33\textwidth}
         \centering
         \scalebox{-1}[1]{\includegraphics[width=0.33\textwidth]{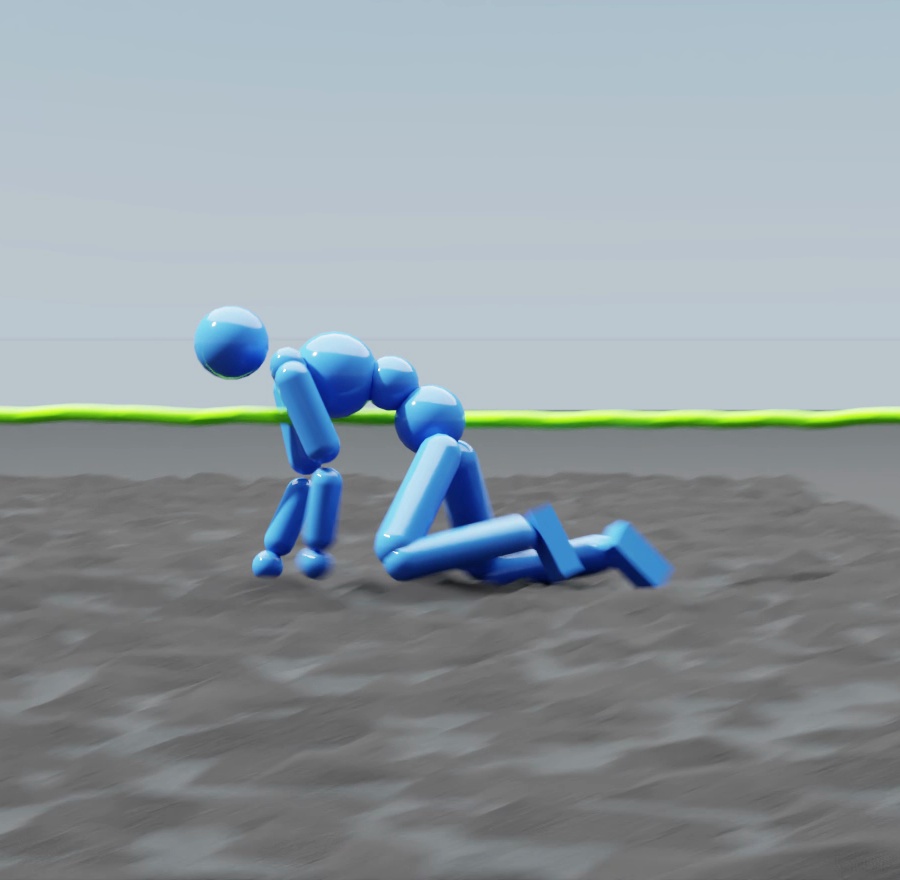}}\hfill
         \scalebox{-1}[1]{\includegraphics[width=0.33\textwidth]{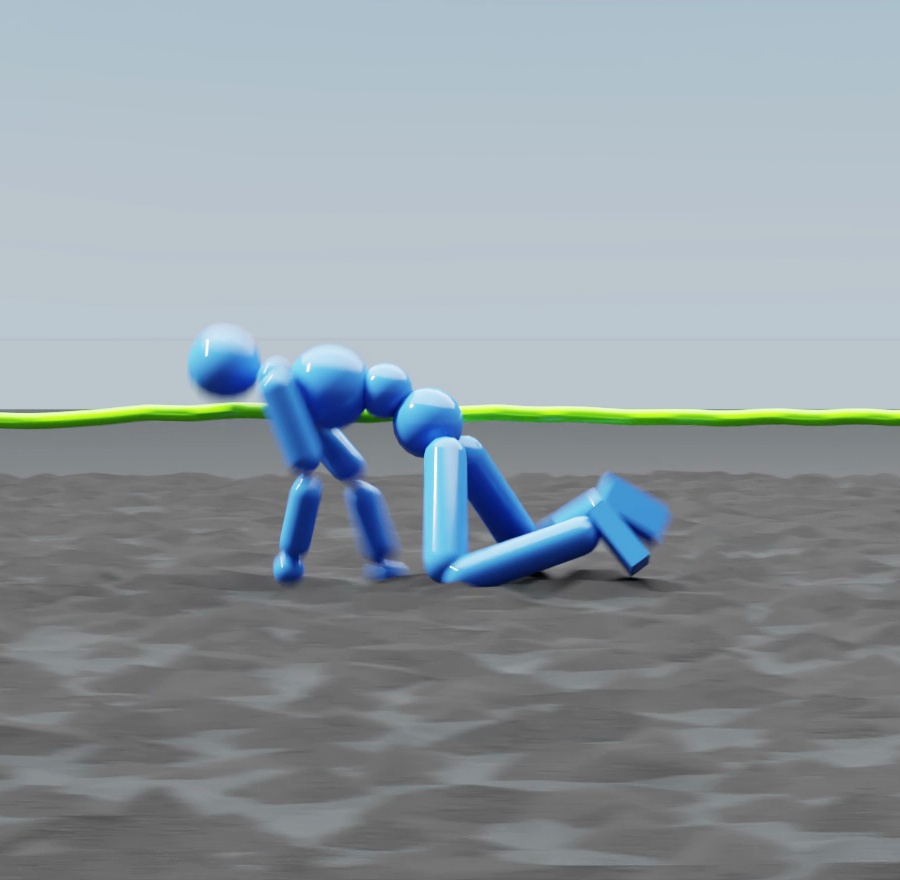}}\hfill
         \scalebox{-1}[1]{\includegraphics[width=0.33\textwidth]{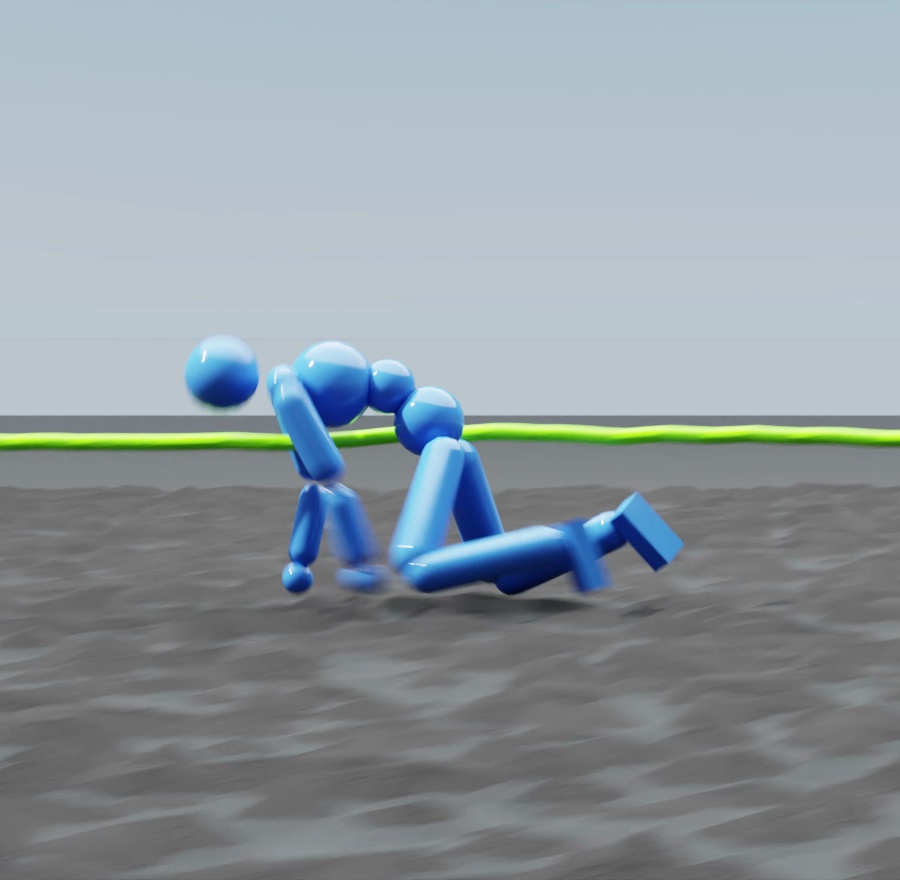}}
         \caption{Crawling on gravel}
         \label{fig: locomotion crawl gravel}
     \end{subfigure}\hfill
     \begin{subfigure}[b]{0.33\textwidth}
         \centering
         \scalebox{-1}[1]{\includegraphics[width=0.33\textwidth]{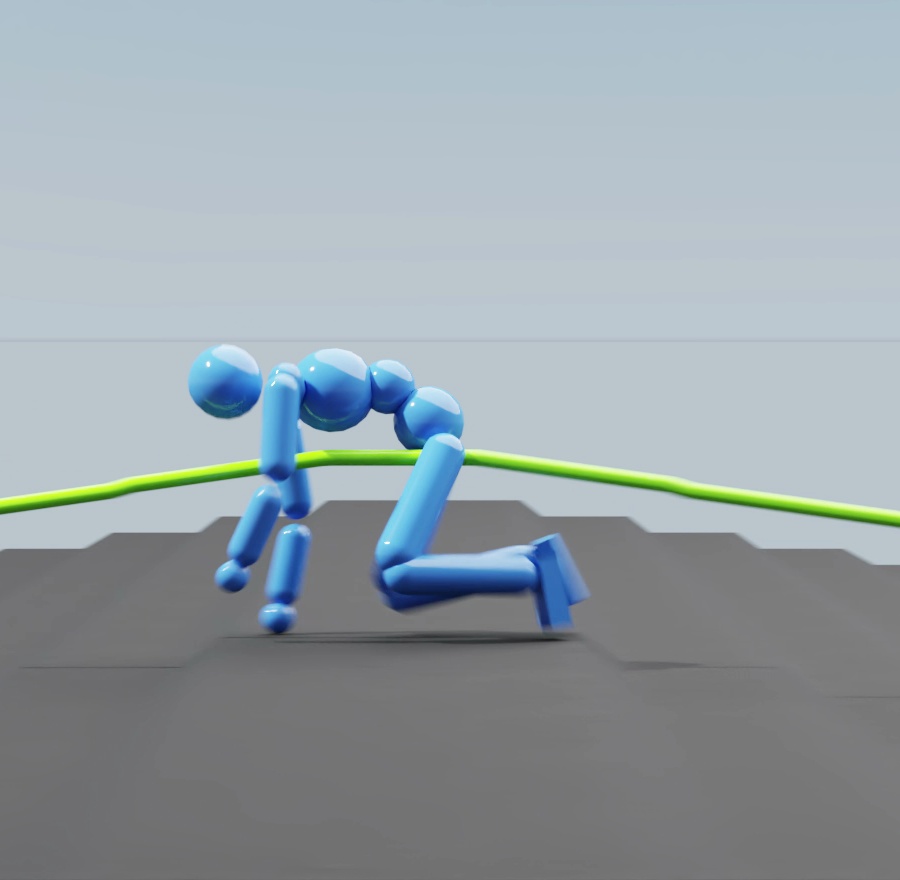}}\hfill
         \scalebox{-1}[1]{\includegraphics[width=0.33\textwidth]{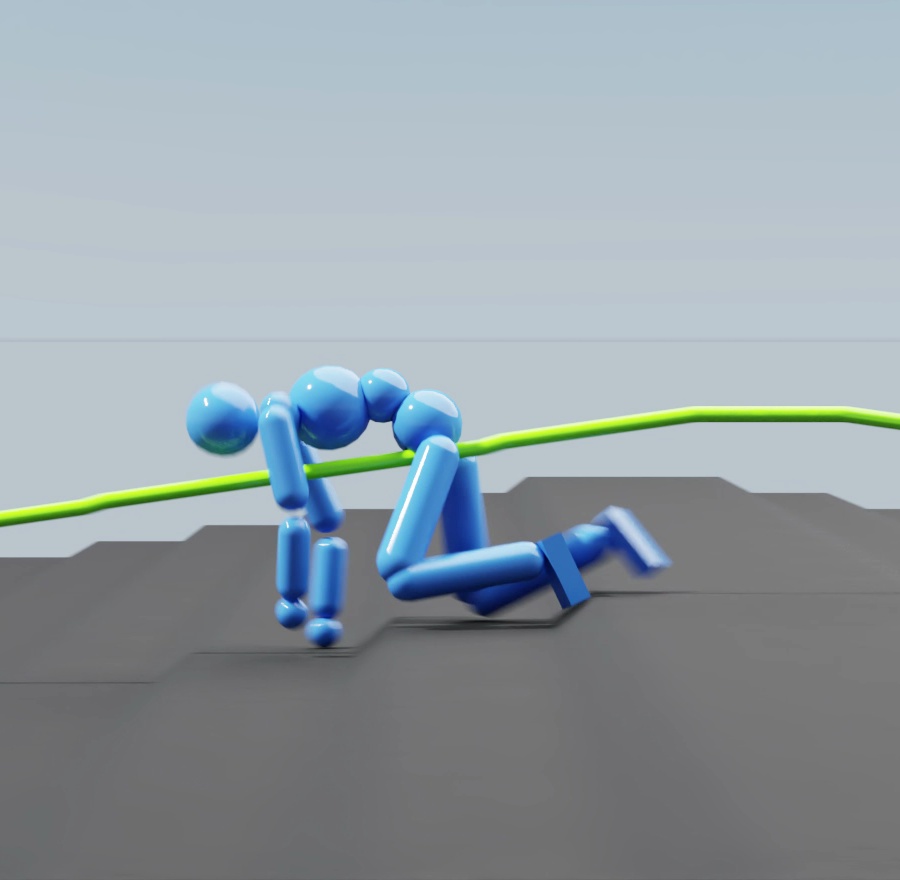}}\hfill
         \scalebox{-1}[1]{\includegraphics[width=0.33\textwidth]{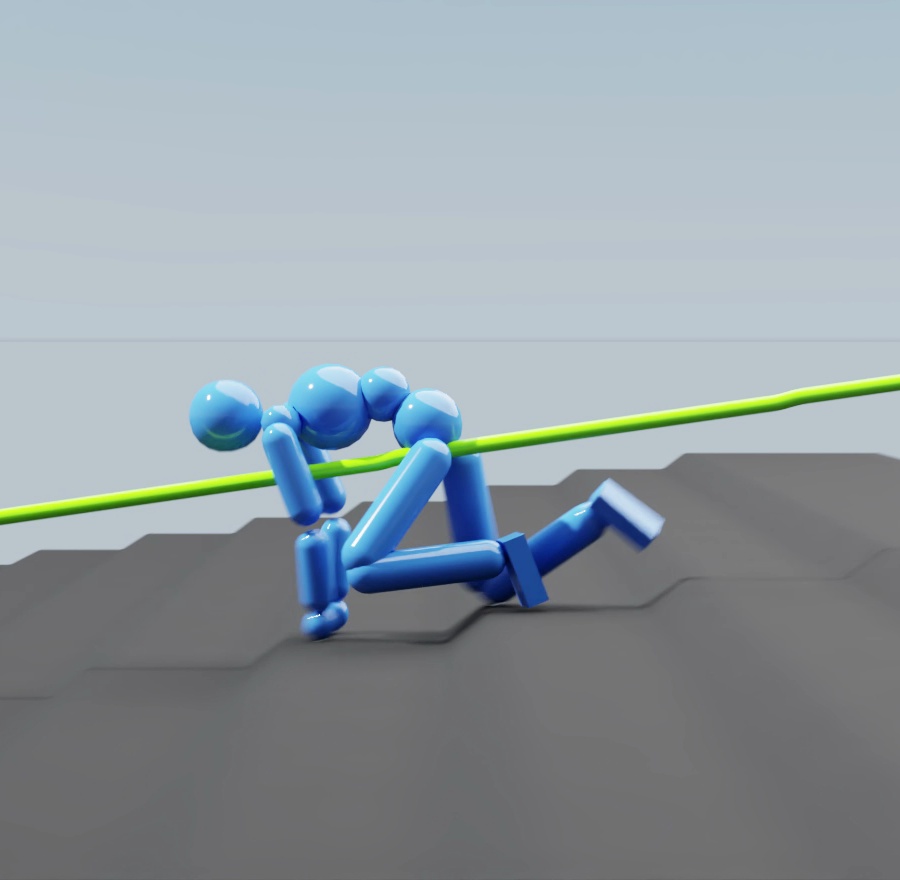}}
         \caption{Crawling down stairs}
         \label{fig: locomotion crawl stairs}
     \end{subfigure} \\
     \begin{subfigure}[b]{0.33\textwidth}
         \centering
         \scalebox{-1}[1]{\includegraphics[width=0.33\textwidth]{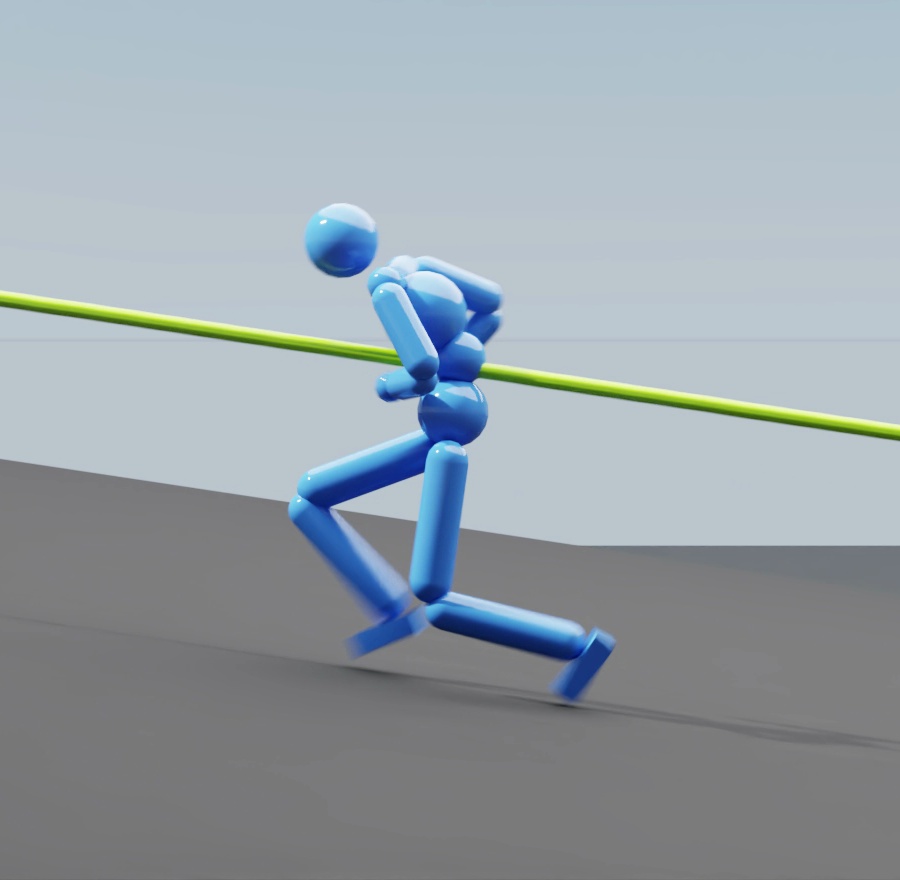}}\hfill
         \scalebox{-1}[1]{\includegraphics[width=0.33\textwidth]{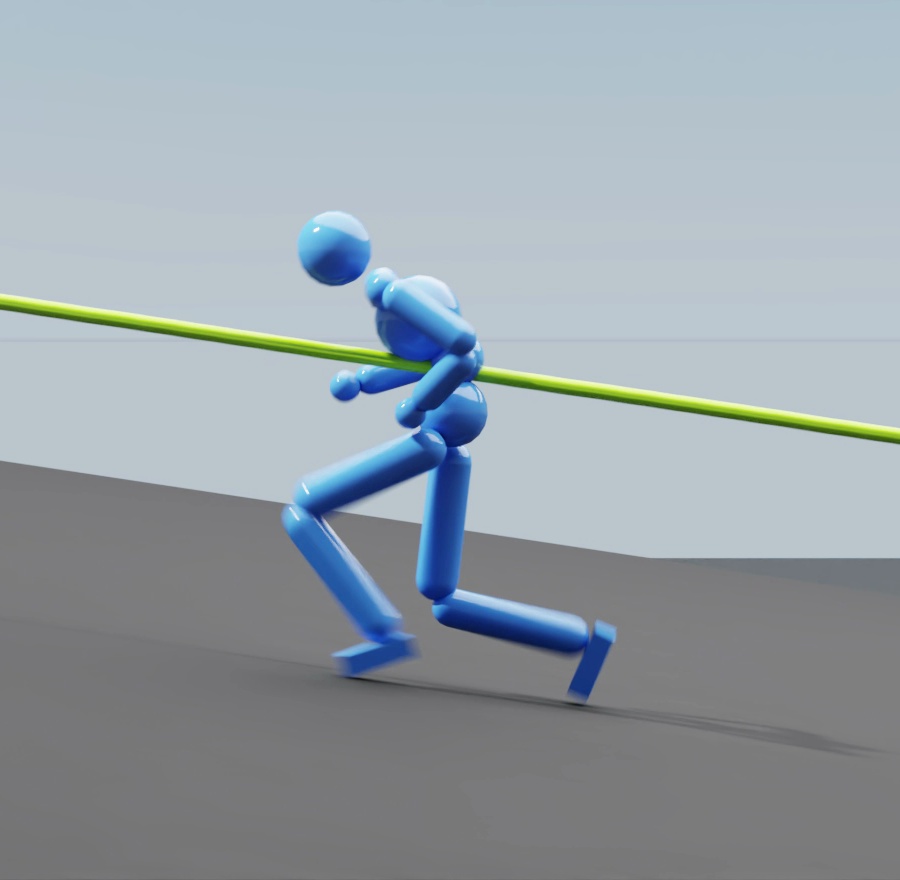}}\hfill
         \scalebox{-1}[1]{\includegraphics[width=0.33\textwidth]{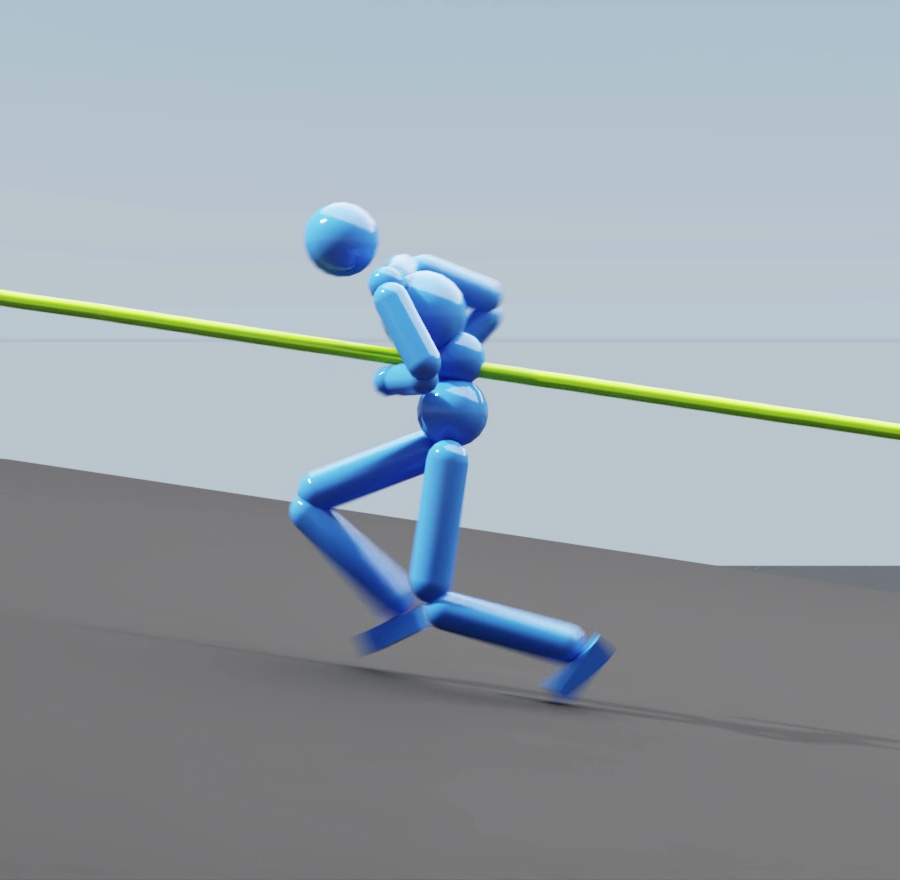}}
         \caption{Crouch-walking up a slope}
         \label{fig: locomotion crouch slope}
     \end{subfigure}\hfill
     \begin{subfigure}[b]{0.33\textwidth}
         \centering
         \scalebox{-1}[1]{\includegraphics[width=0.33\textwidth]{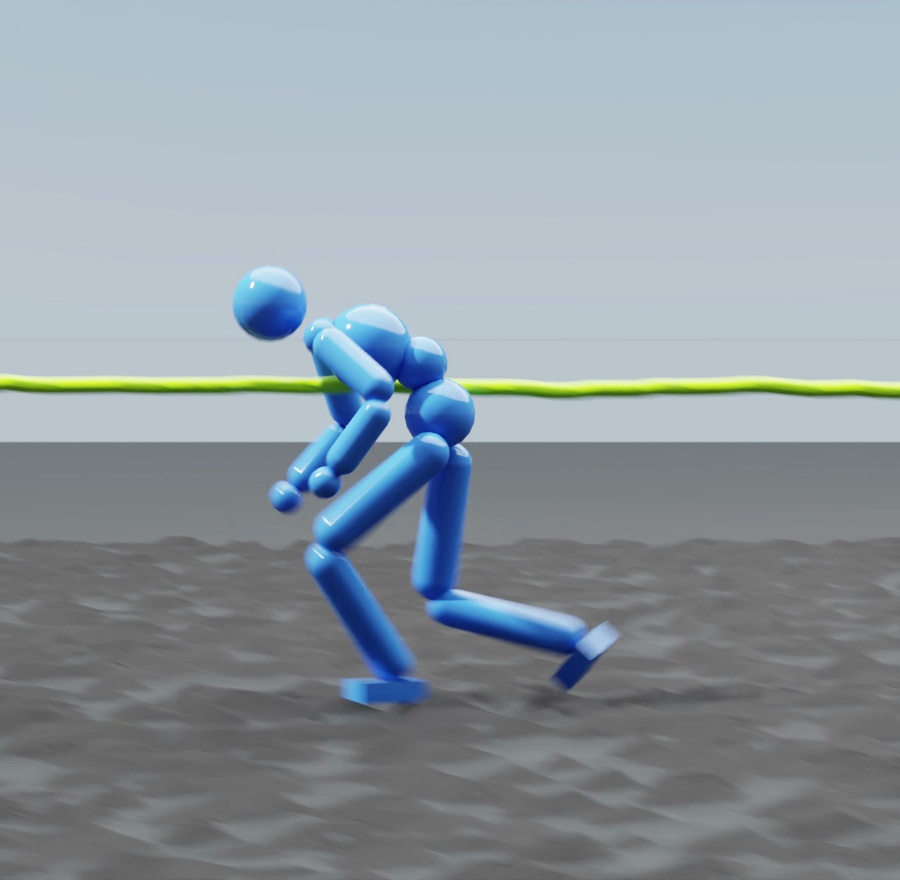}}\hfill
         \scalebox{-1}[1]{\includegraphics[width=0.33\textwidth]{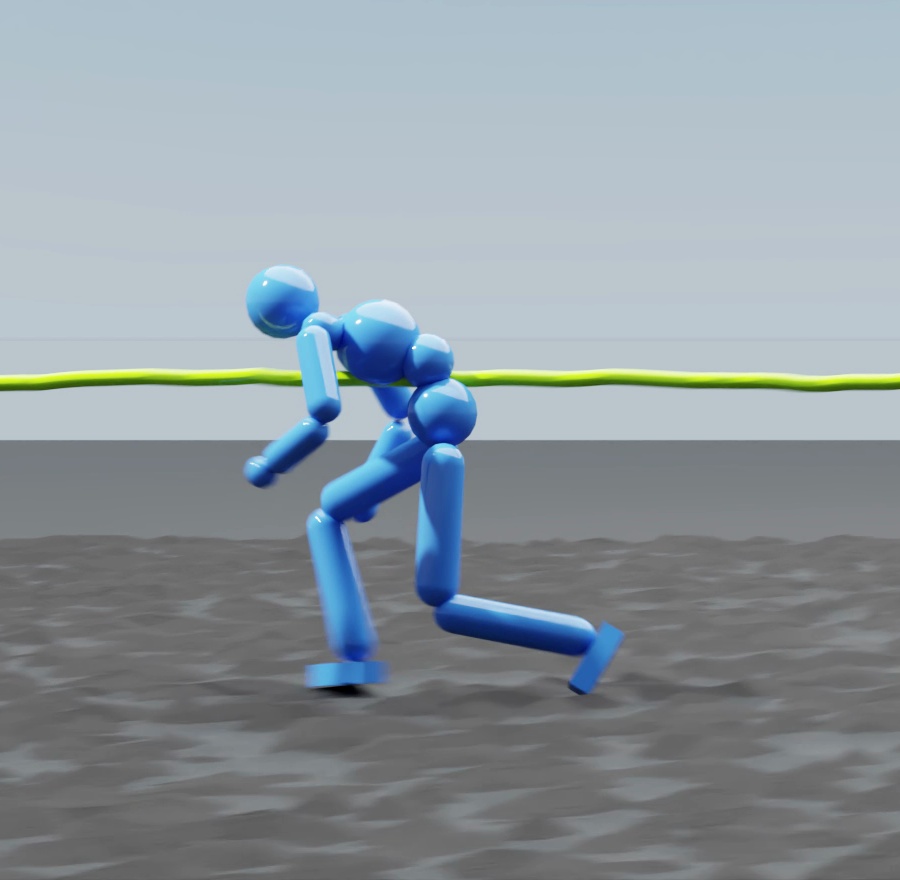}}\hfill
         \scalebox{-1}[1]{\includegraphics[width=0.33\textwidth]{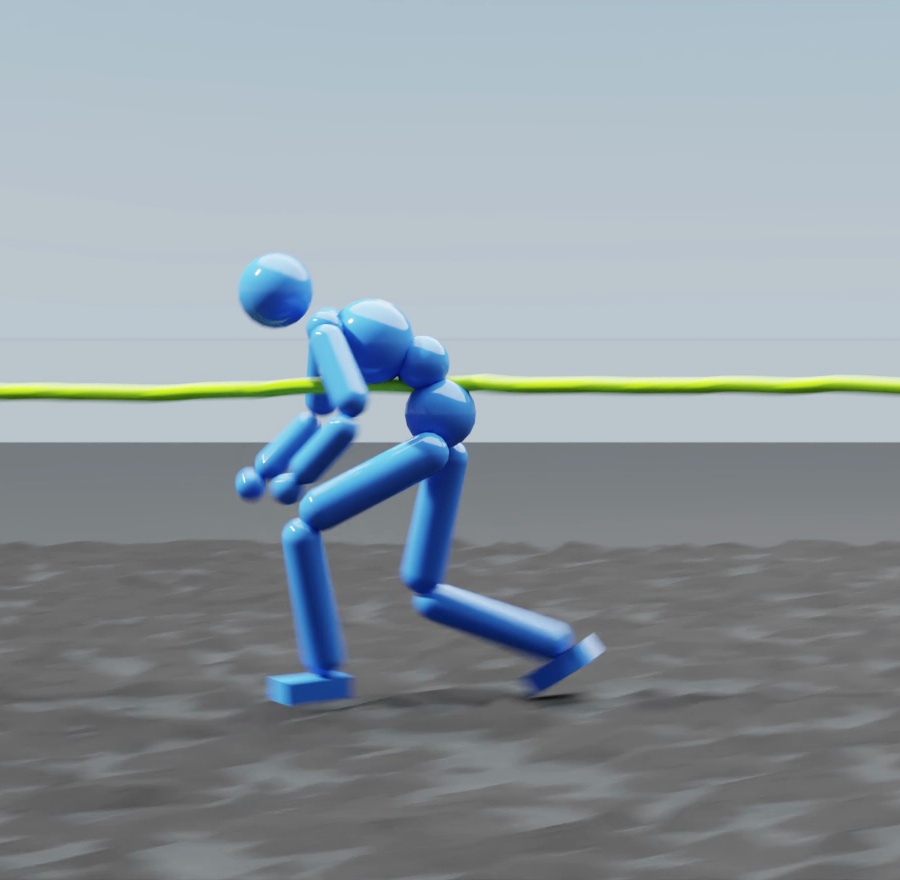}}
         \caption{Crouch-walking on rough terrain}
         \label{fig: locomotion crouch gravel}
     \end{subfigure}\hfill
     \begin{subfigure}[b]{0.33\textwidth}
         \centering
         \scalebox{-1}[1]{\includegraphics[width=0.33\textwidth]{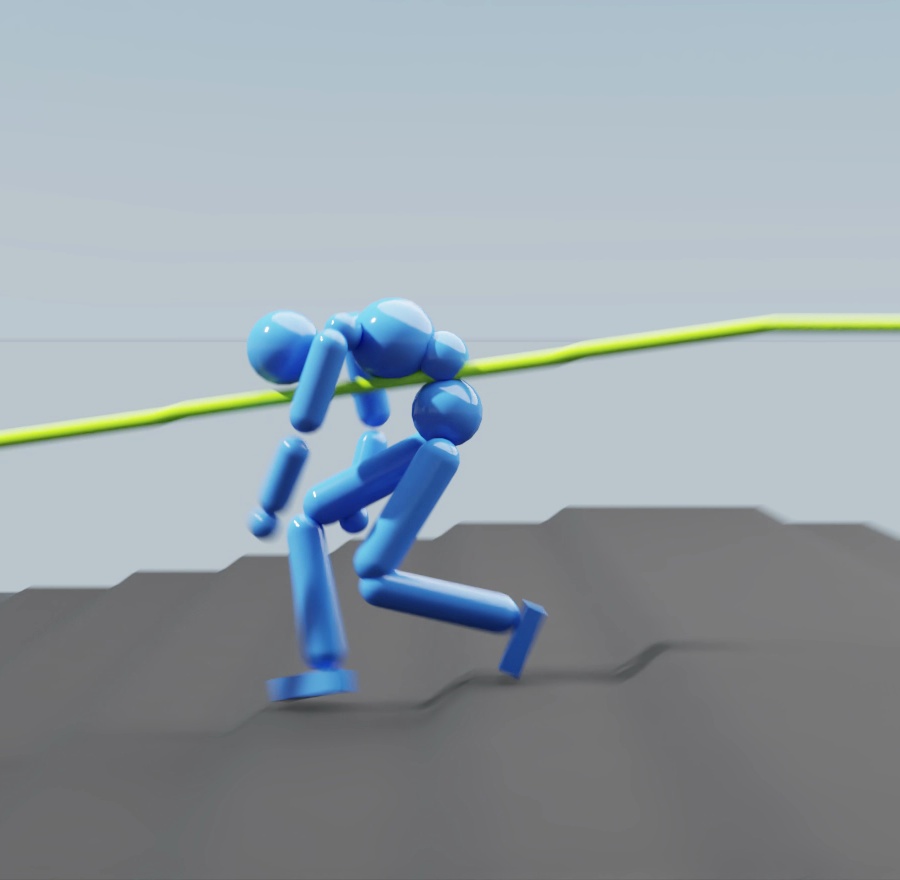}}\hfill
         \scalebox{-1}[1]{\includegraphics[width=0.33\textwidth]{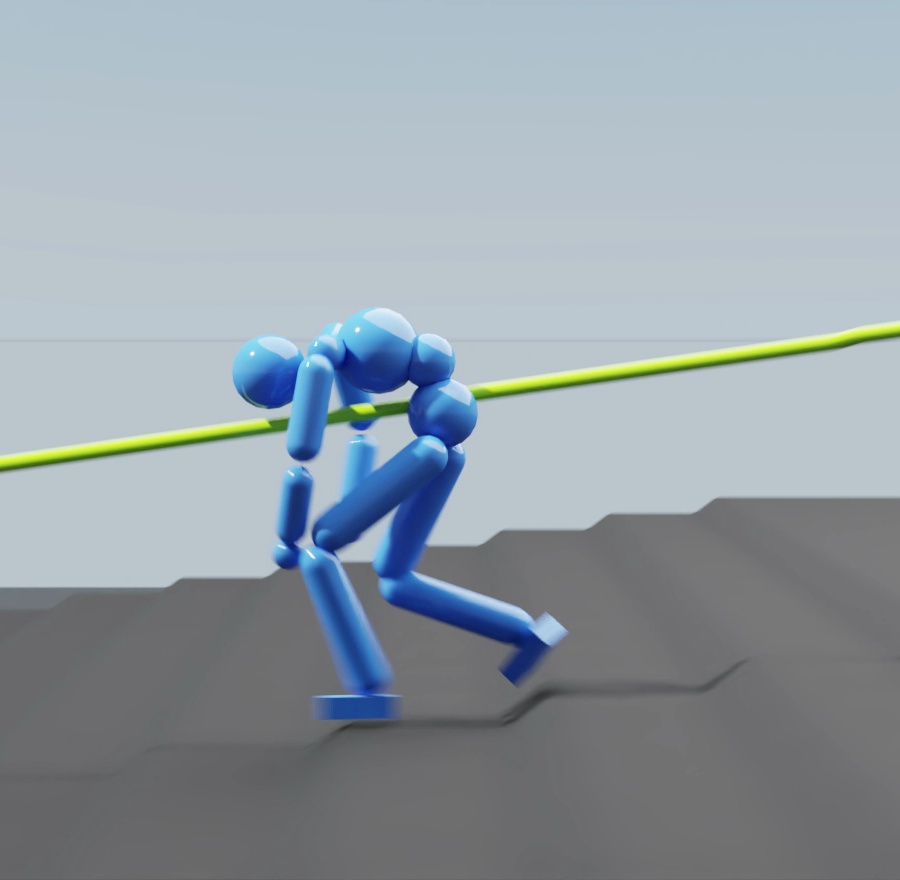}}\hfill
         \scalebox{-1}[1]{\includegraphics[width=0.33\textwidth]{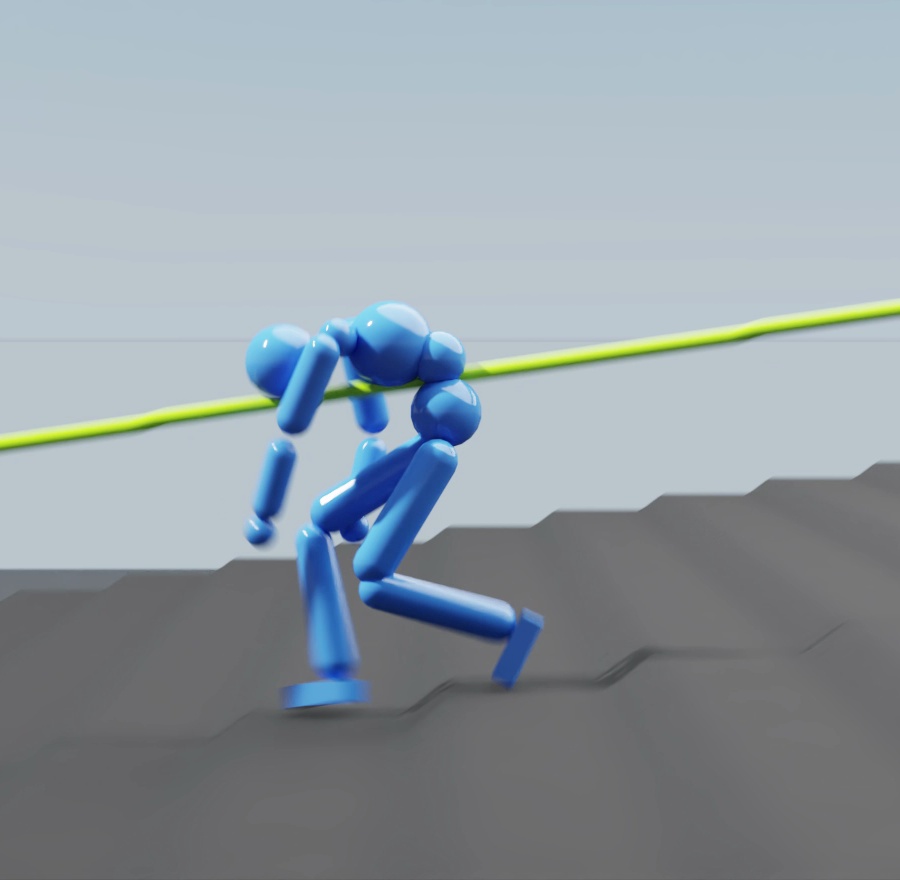}}
         \caption{Crouch-walking down stairs}
         \label{fig: locomotion crouch stairs}
     \end{subfigure} \\
          \begin{subfigure}[b]{0.33\textwidth}
         \centering
         \scalebox{-1}[1]{\includegraphics[width=0.33\textwidth]{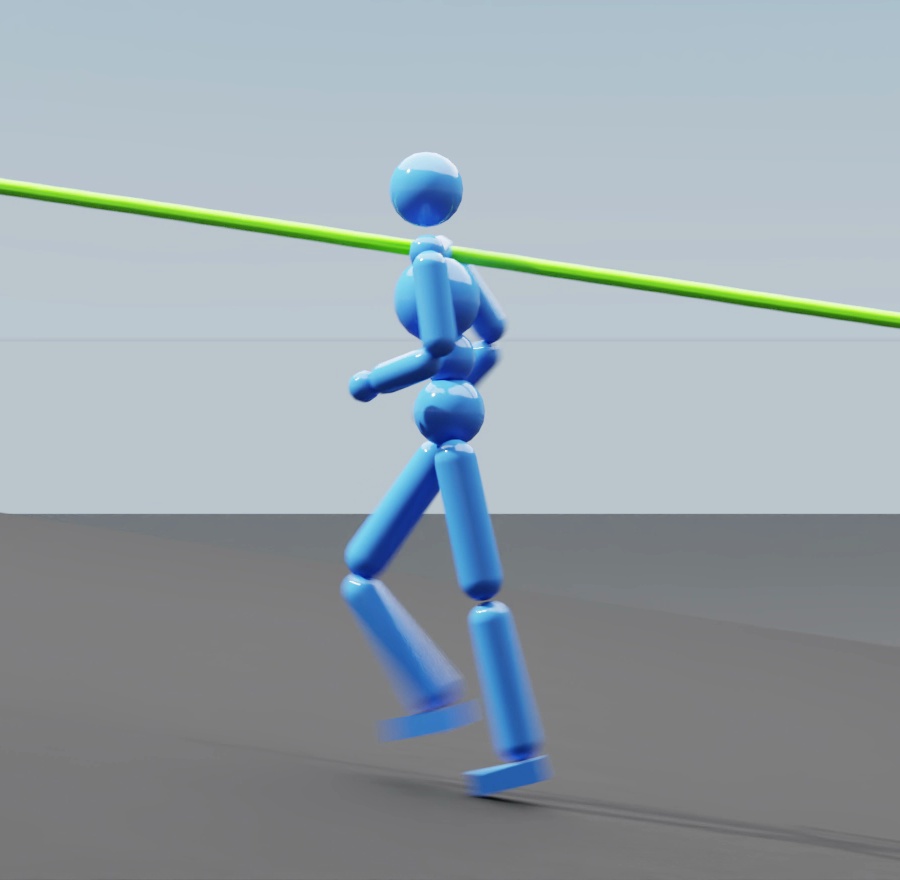}}\hfill
         \scalebox{-1}[1]{\includegraphics[width=0.33\textwidth]{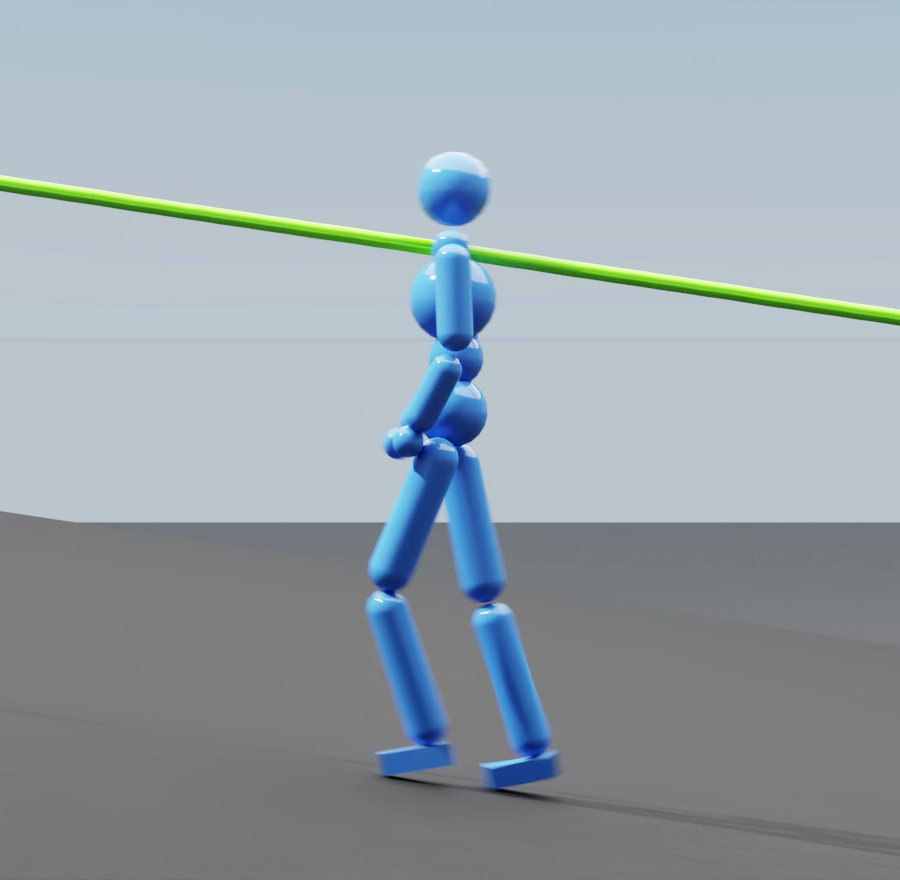}}\hfill
         \scalebox{-1}[1]{\includegraphics[width=0.33\textwidth]{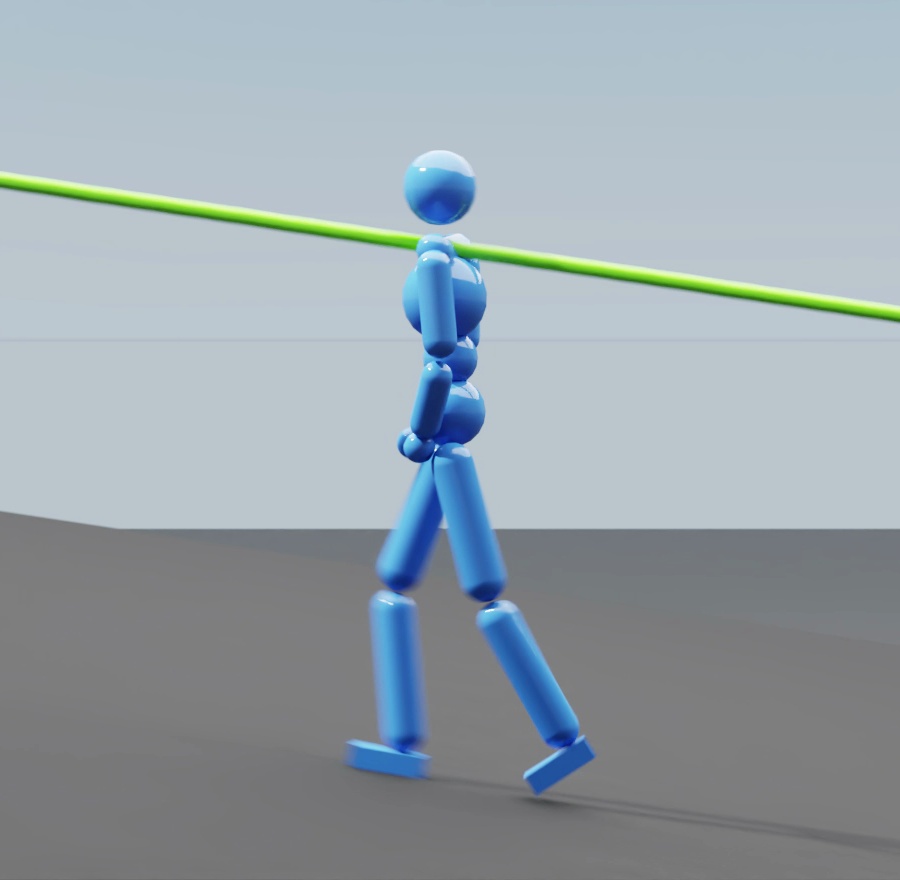}}
         \caption{Walking up a slope}
         \label{fig: locomotion walk slope}
     \end{subfigure}\hfill
     \begin{subfigure}[b]{0.33\textwidth}
         \centering
         \scalebox{-1}[1]{\includegraphics[width=0.33\textwidth]{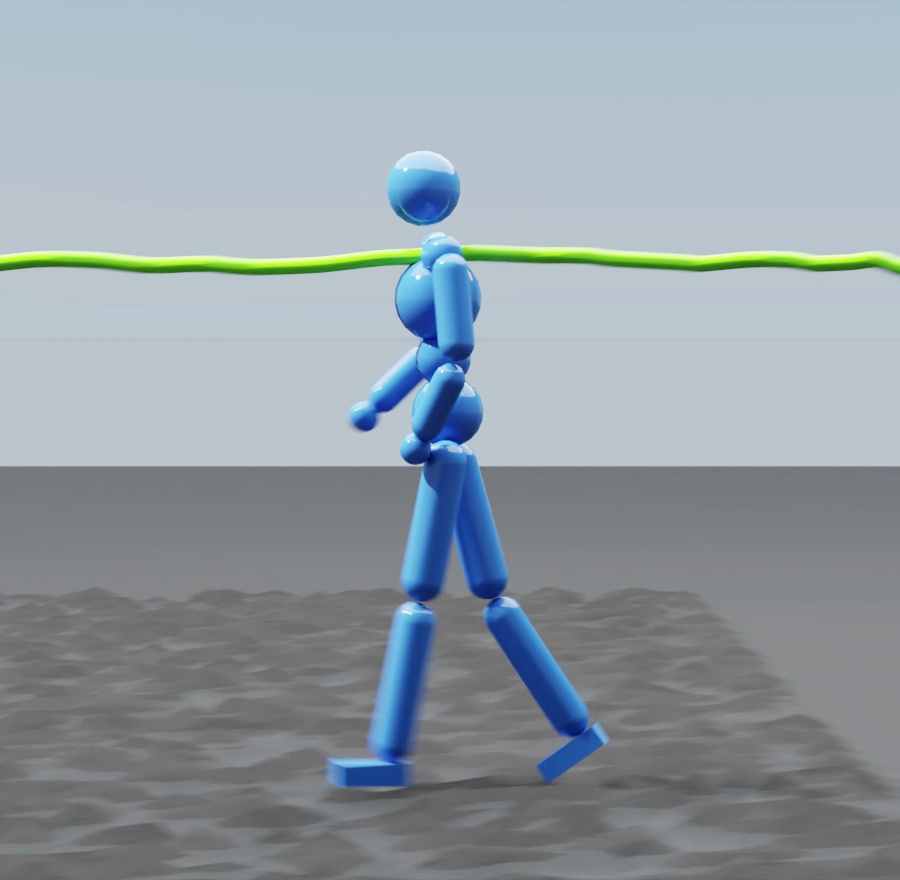}}\hfill
         \scalebox{-1}[1]{\includegraphics[width=0.33\textwidth]{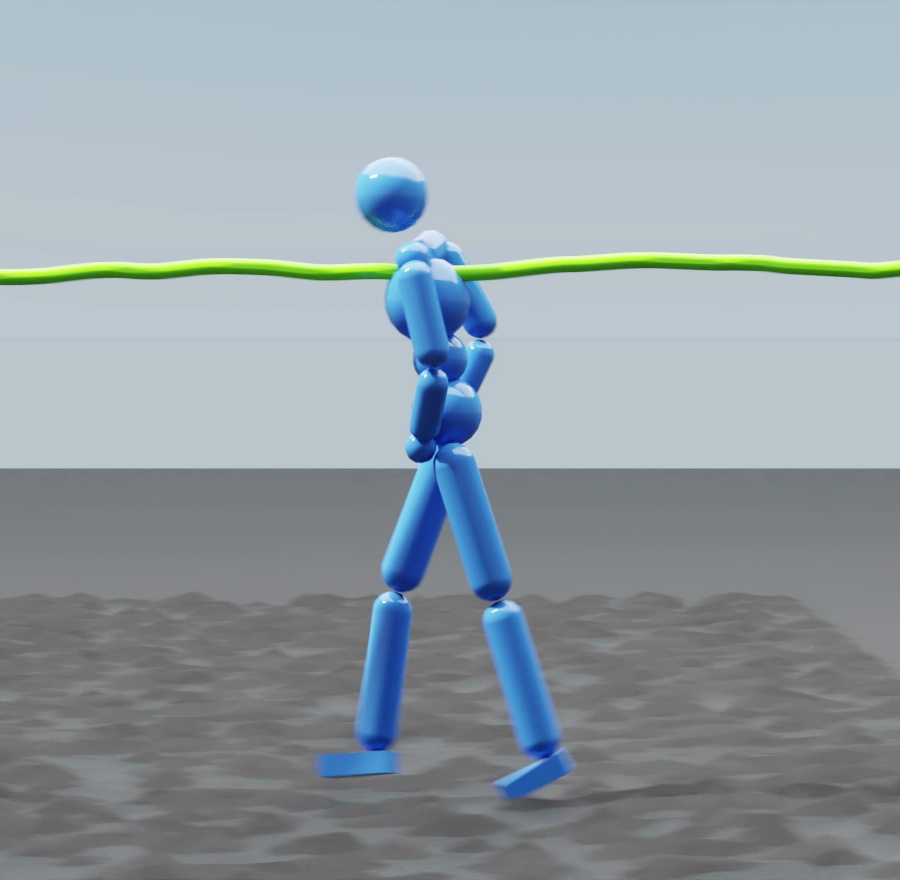}}\hfill
         \scalebox{-1}[1]{\includegraphics[width=0.33\textwidth]{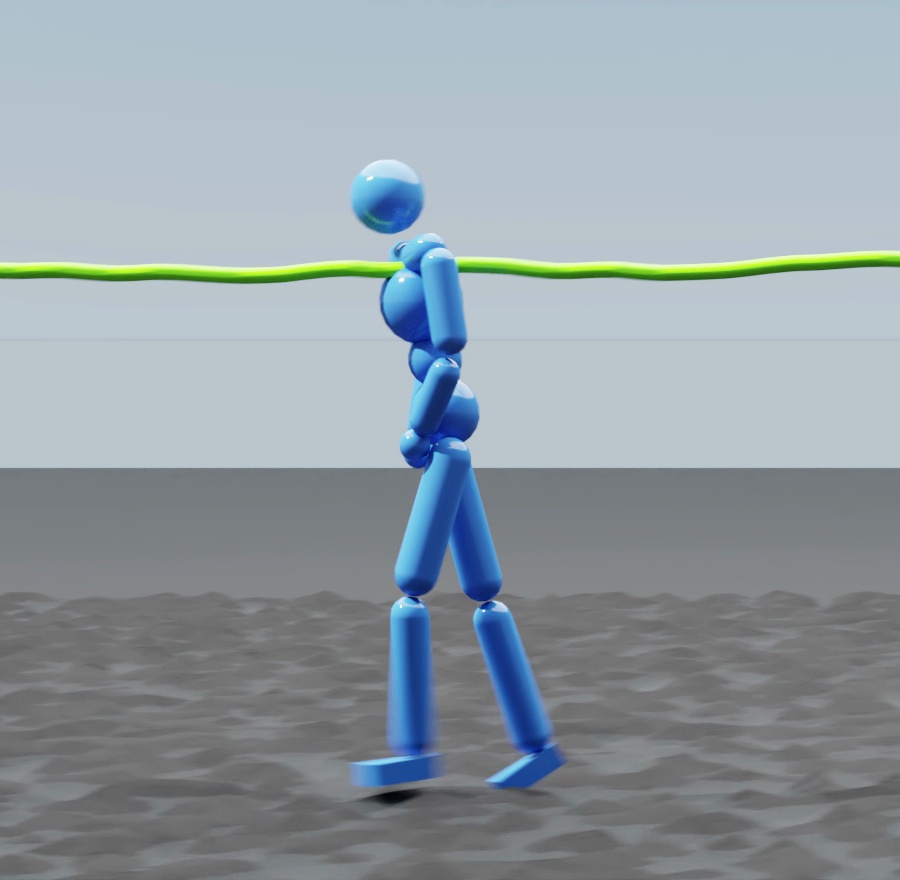}}
         \caption{Walking on rough terrain}
         \label{fig: locomotion walk gravel}
     \end{subfigure}\hfill
     \begin{subfigure}[b]{0.33\textwidth}
         \centering
         \scalebox{-1}[1]{\includegraphics[width=0.33\textwidth]{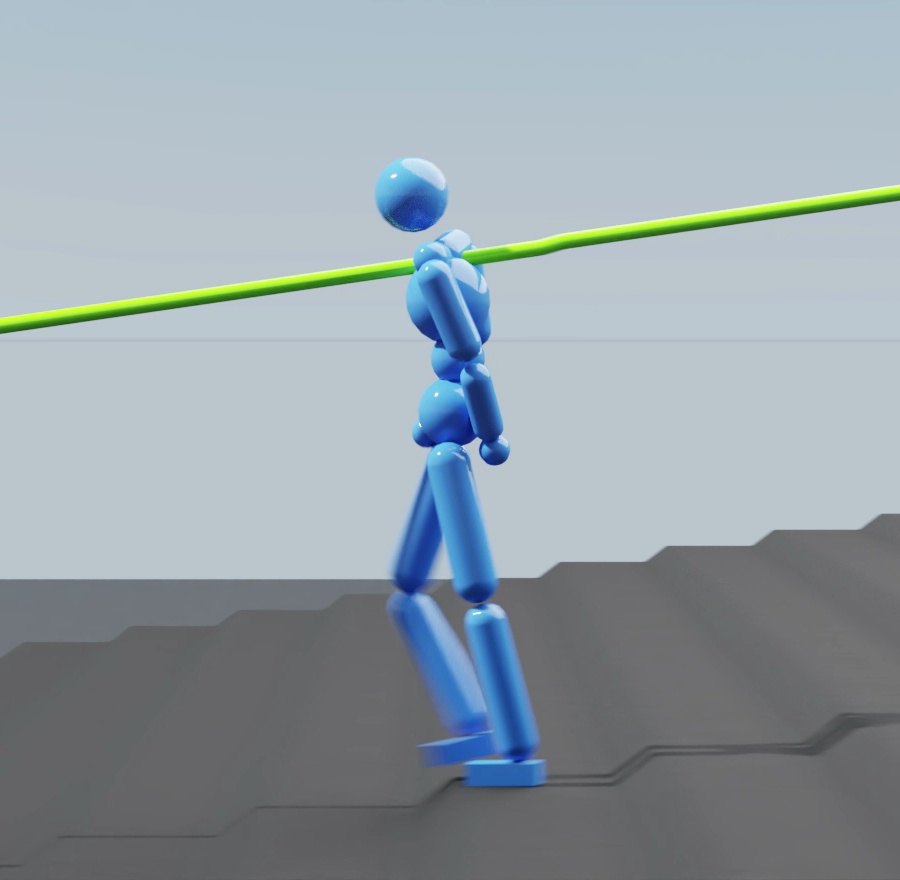}}\hfill
         \scalebox{-1}[1]{\includegraphics[width=0.33\textwidth]{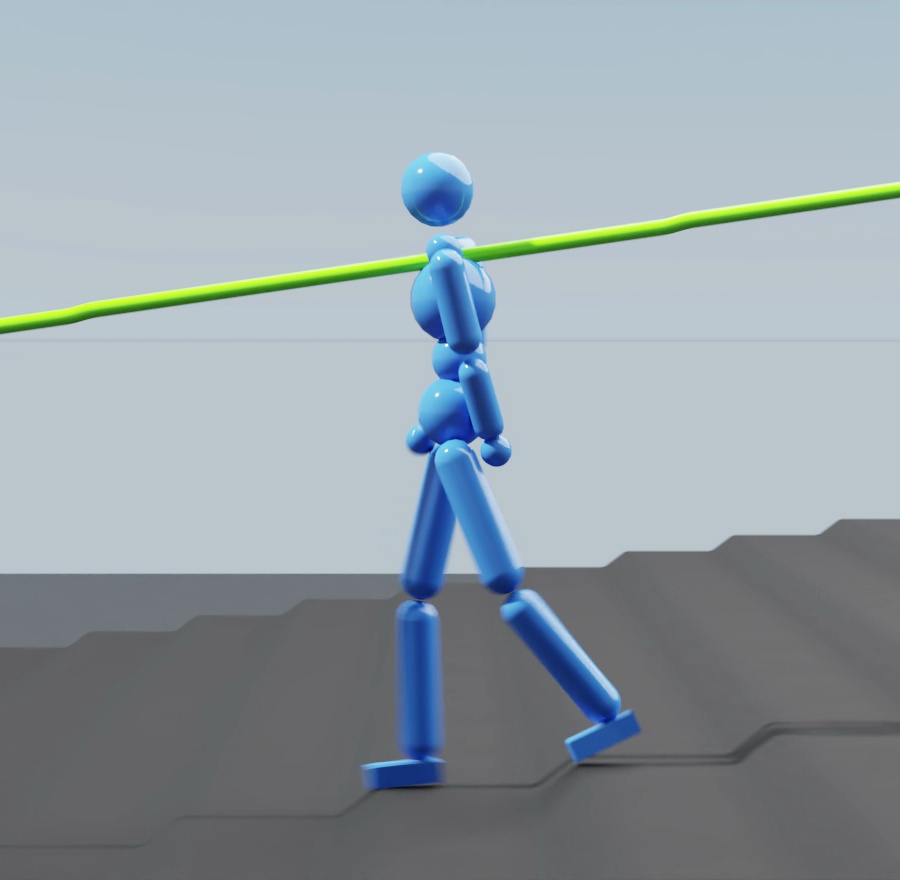}}\hfill
         \scalebox{-1}[1]{\includegraphics[width=0.33\textwidth]{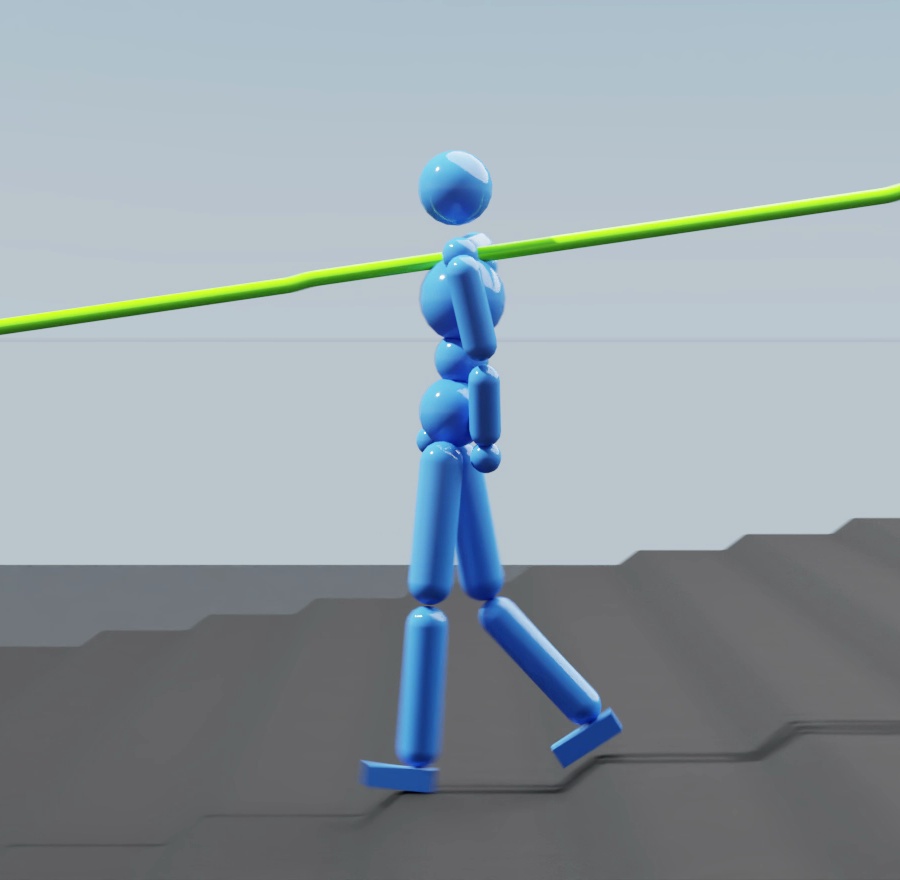}}
         \caption{Walking down stairs}
         \label{fig: locomotion walk stairs}
     \end{subfigure} \\
     \begin{subfigure}[b]{0.33\textwidth}
         \centering
         \scalebox{-1}[1]{\includegraphics[width=0.33\textwidth]{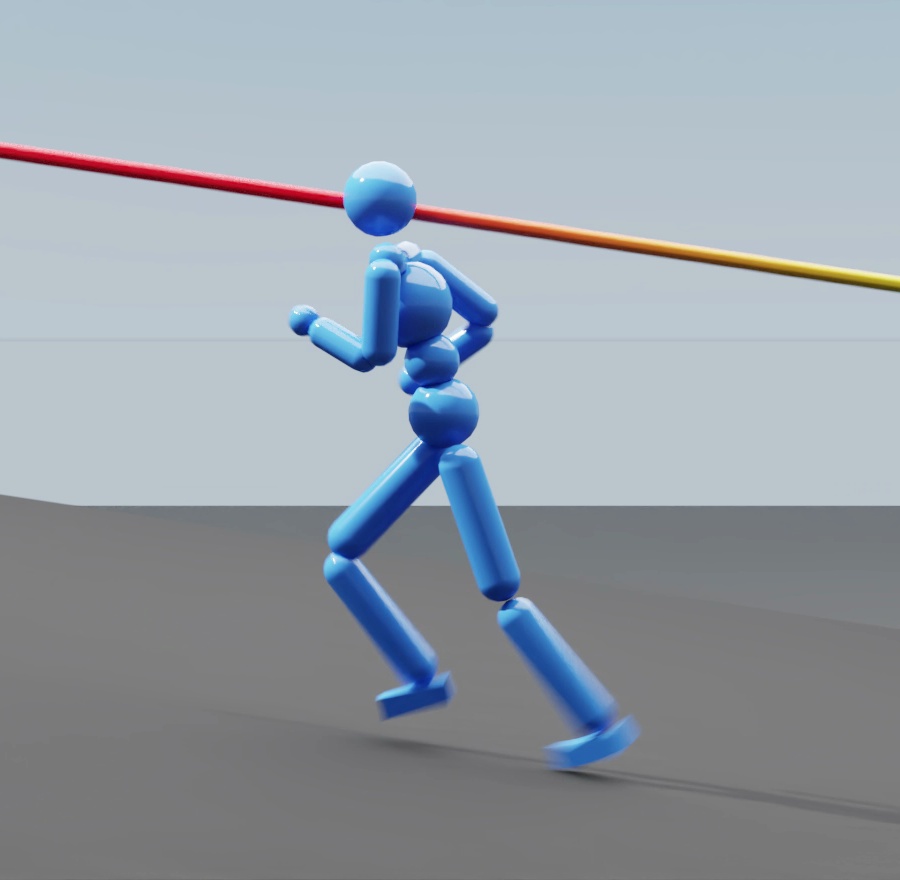}}\hfill
         \scalebox{-1}[1]{\includegraphics[width=0.33\textwidth]{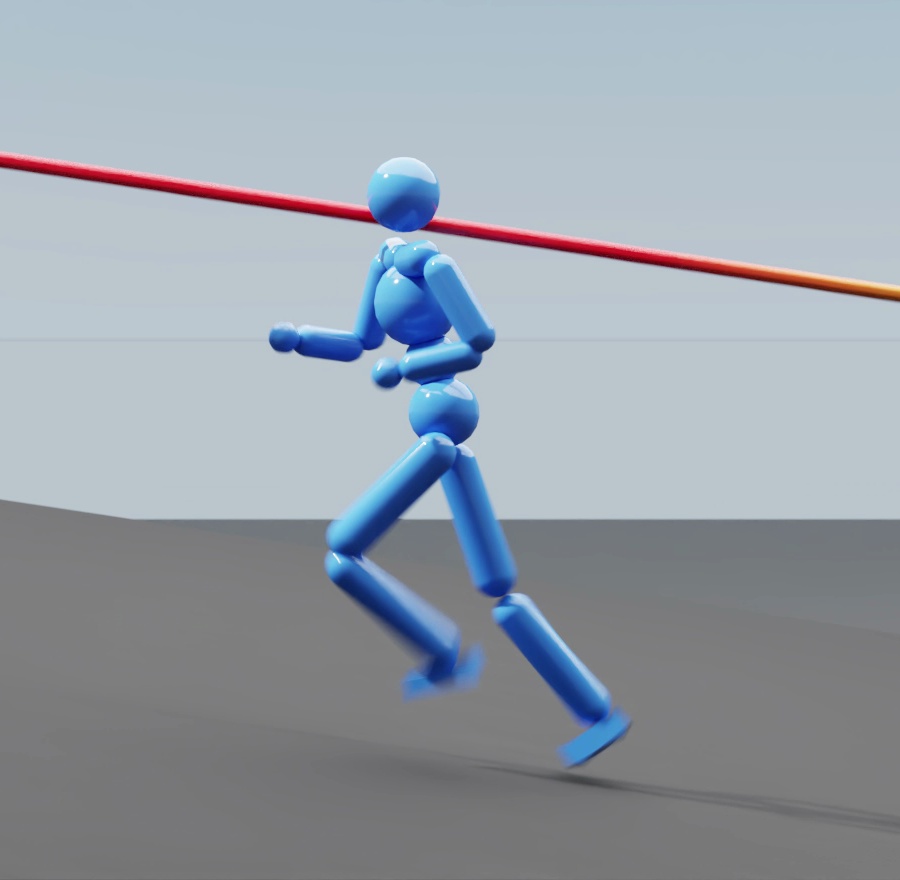}}\hfill
         \scalebox{-1}[1]{\includegraphics[width=0.33\textwidth]{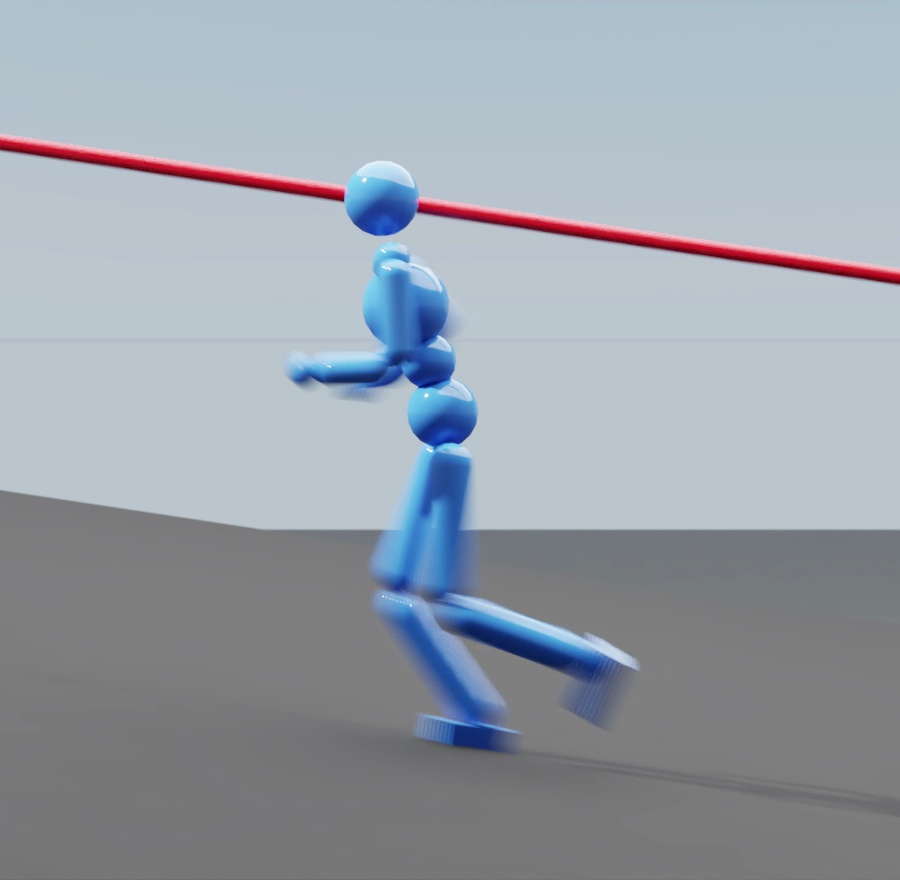}}
         \caption{Running up a slope}
         \label{fig: locomotion run slope}
     \end{subfigure}
     \begin{subfigure}[b]{0.33\textwidth}
         \centering
         \scalebox{-1}[1]{\includegraphics[width=0.33\textwidth]{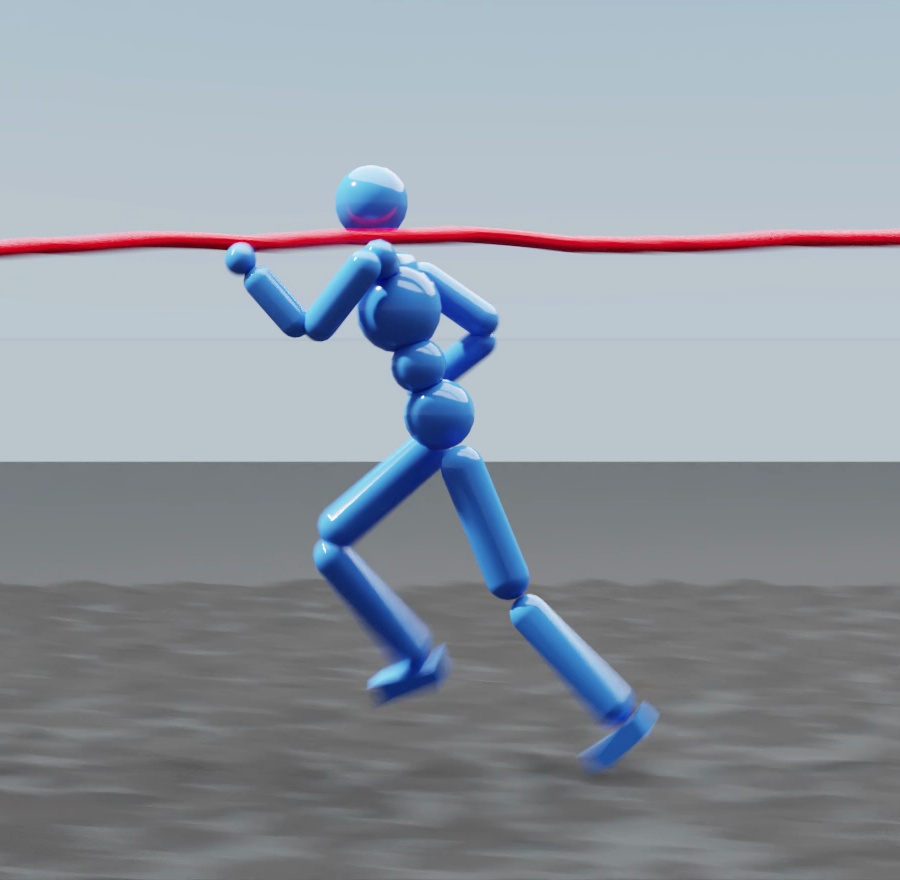}}\hfill
         \scalebox{-1}[1]{\includegraphics[width=0.33\textwidth]{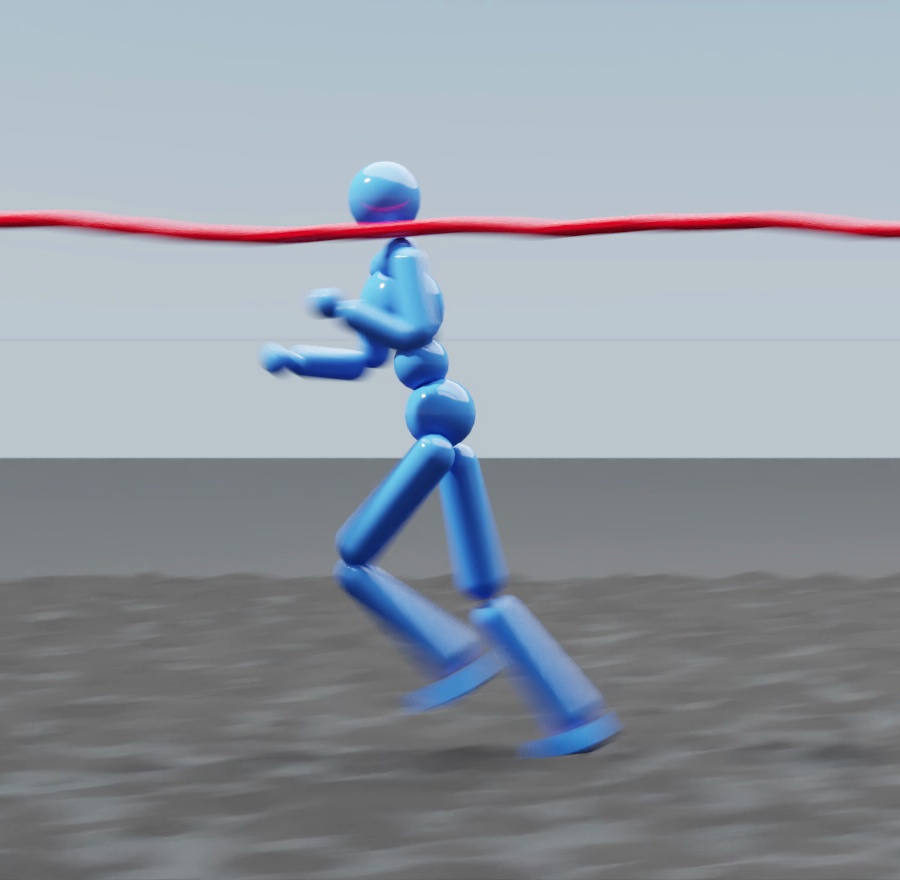}}\hfill
         \scalebox{-1}[1]{\includegraphics[width=0.33\textwidth]{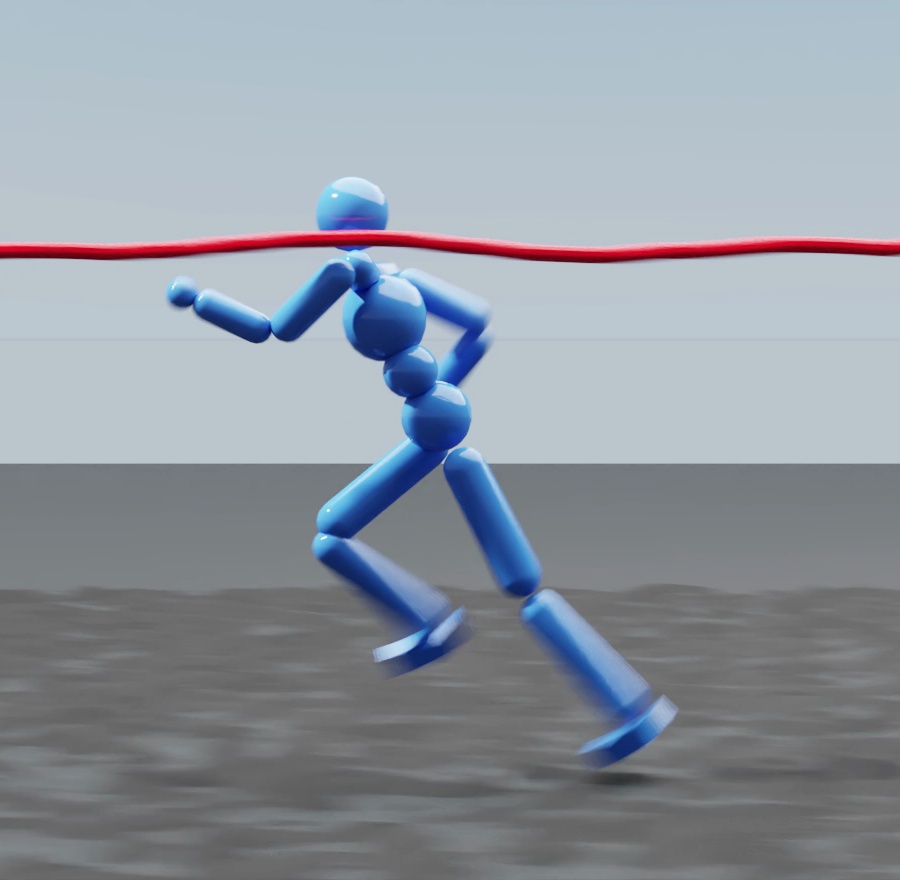}}
         \caption{Running on rough terrain}
         \label{fig: locomotion run gravel}
     \end{subfigure}\hfill
     \begin{subfigure}[b]{0.33\textwidth}
         \centering
         \scalebox{-1}[1]{\includegraphics[width=0.33\textwidth]{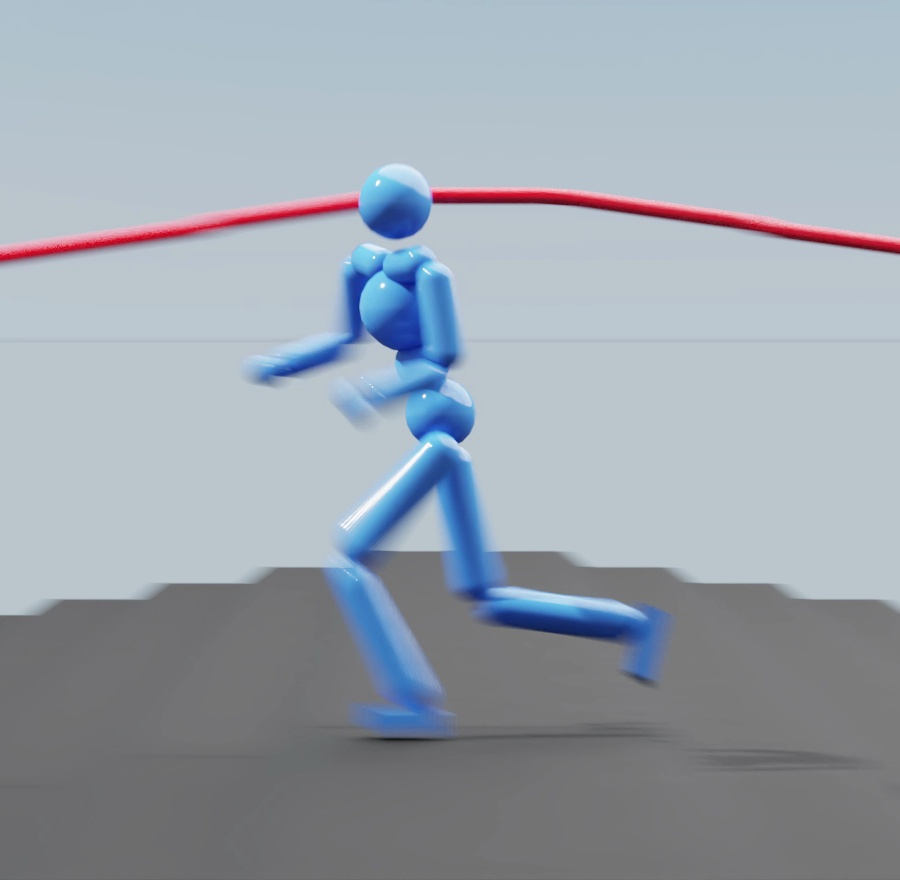}}\hfill
         \scalebox{-1}[1]{\includegraphics[width=0.33\textwidth]{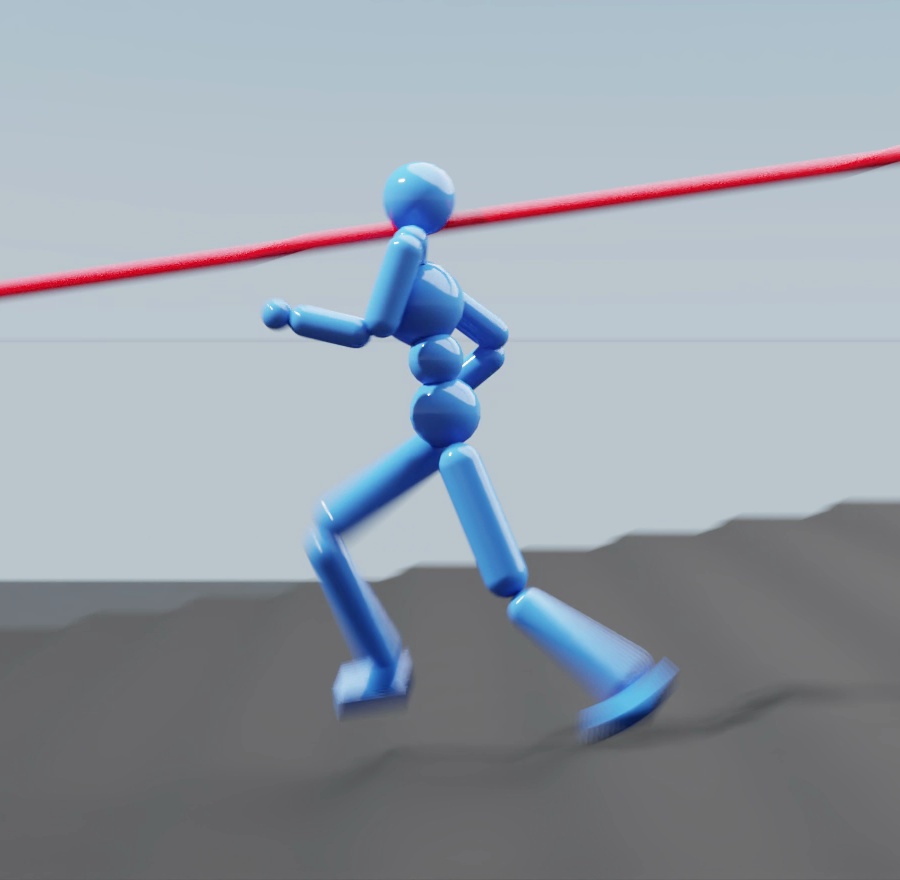}}\hfill
         \scalebox{-1}[1]{\includegraphics[width=0.33\textwidth]{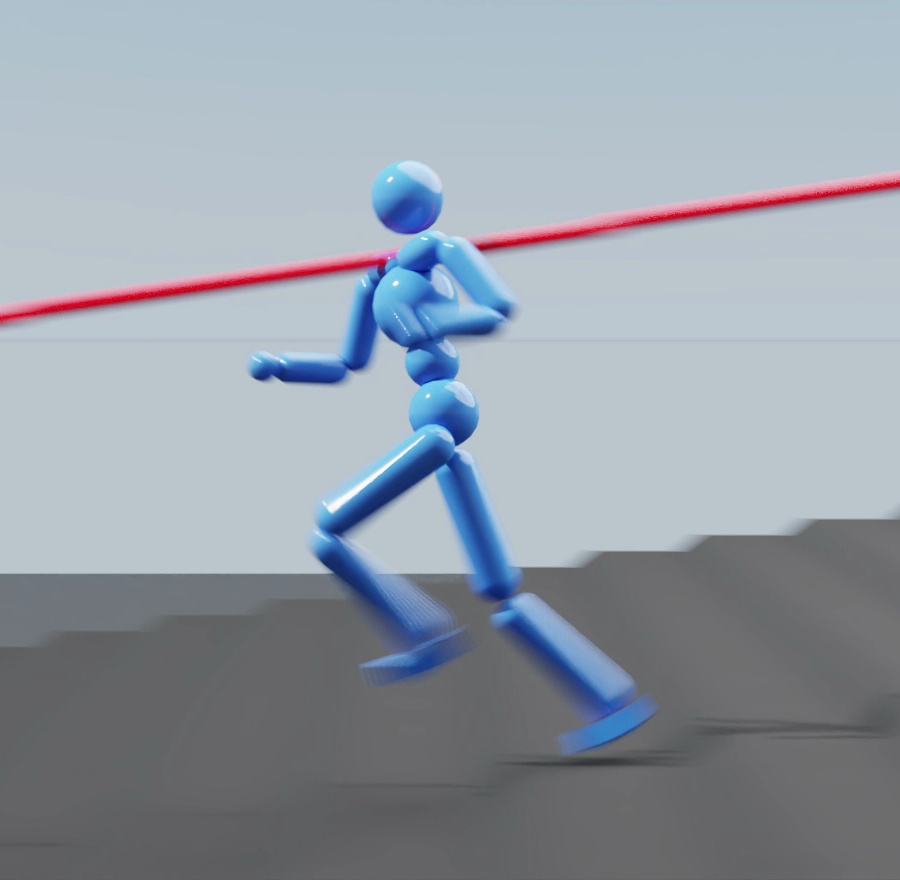}}
         \caption{Running down stairs}
         \label{fig: locomotion run stairs}
     \end{subfigure}
     \caption{\textbf{Locomotion:} Screen captures of our trained controller when conditioned on various forms of locomotion across varying types of terrains. Crawling is achieved by requesting a head-height of $0.4 \left[m\right]$, crouching at $0.8 \left[m\right]$, walking and running at upright $1.47 \left[m\right]$. The corresponding speeds are $1 \left[m/s\right]$, $2 \left[m/s\right]$, $2 \left[m/s\right]$, and $4  \left[m/s\right]$.}
    \label{fig: locomotion example}
\end{figure*}

\begin{figure*}[h]
     \centering
     \begin{subfigure}[b]{0.49\textwidth}
         \centering
         \includegraphics[trim={5cm 5cm 0cm 1cm},clip,width=\textwidth]{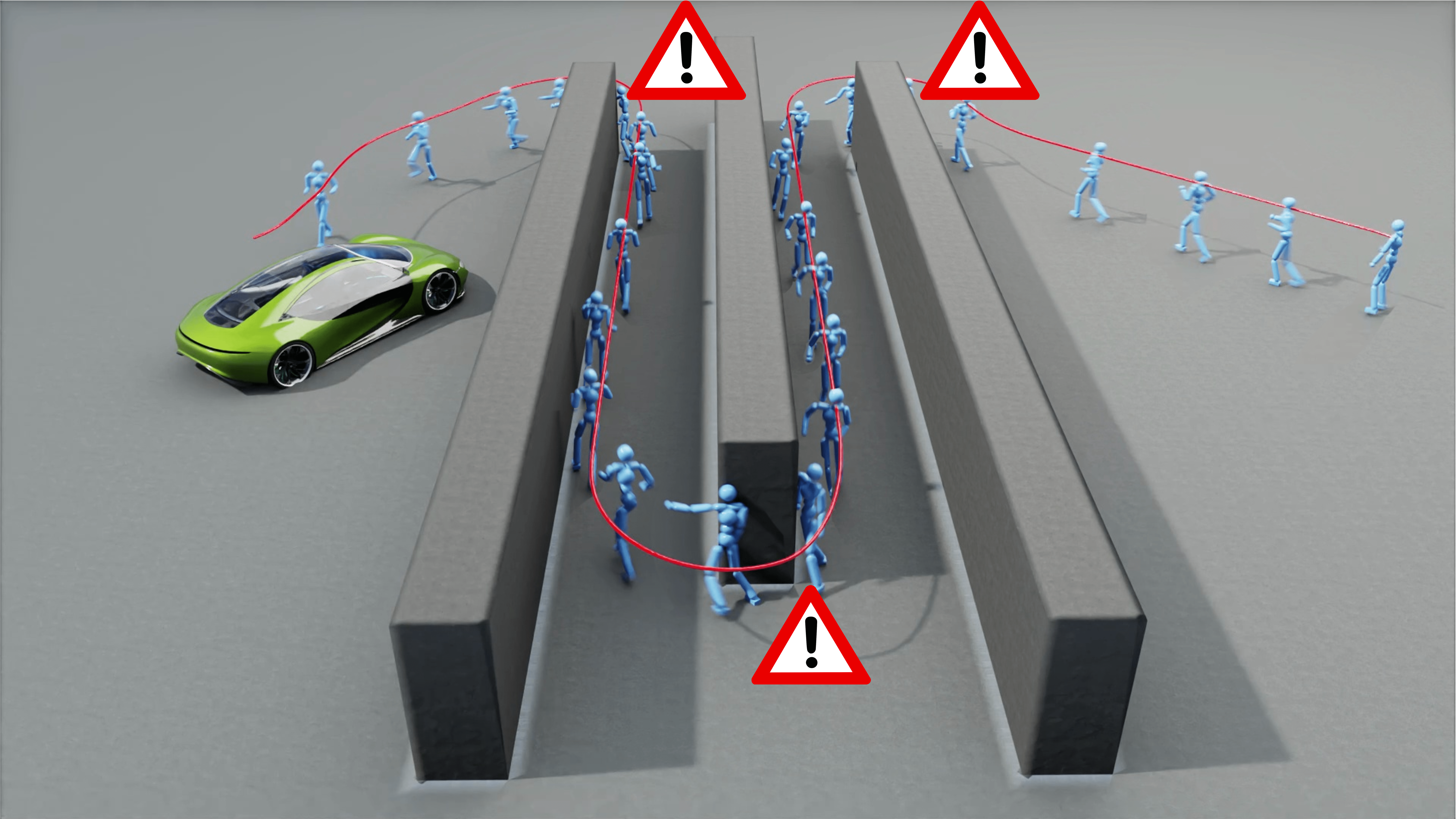}
         \caption{\textbf{Slalom test with constant speed (baseline).} The humanoid follows the path poorly, hitting the walls on every turn.}
         \label{fig: slalom naive}
     \end{subfigure}\hfill
     \begin{subfigure}[b]{0.46\textwidth}
         \centering
         \includegraphics[trim={5cm 5cm 0cm 1cm},clip,width=\textwidth]{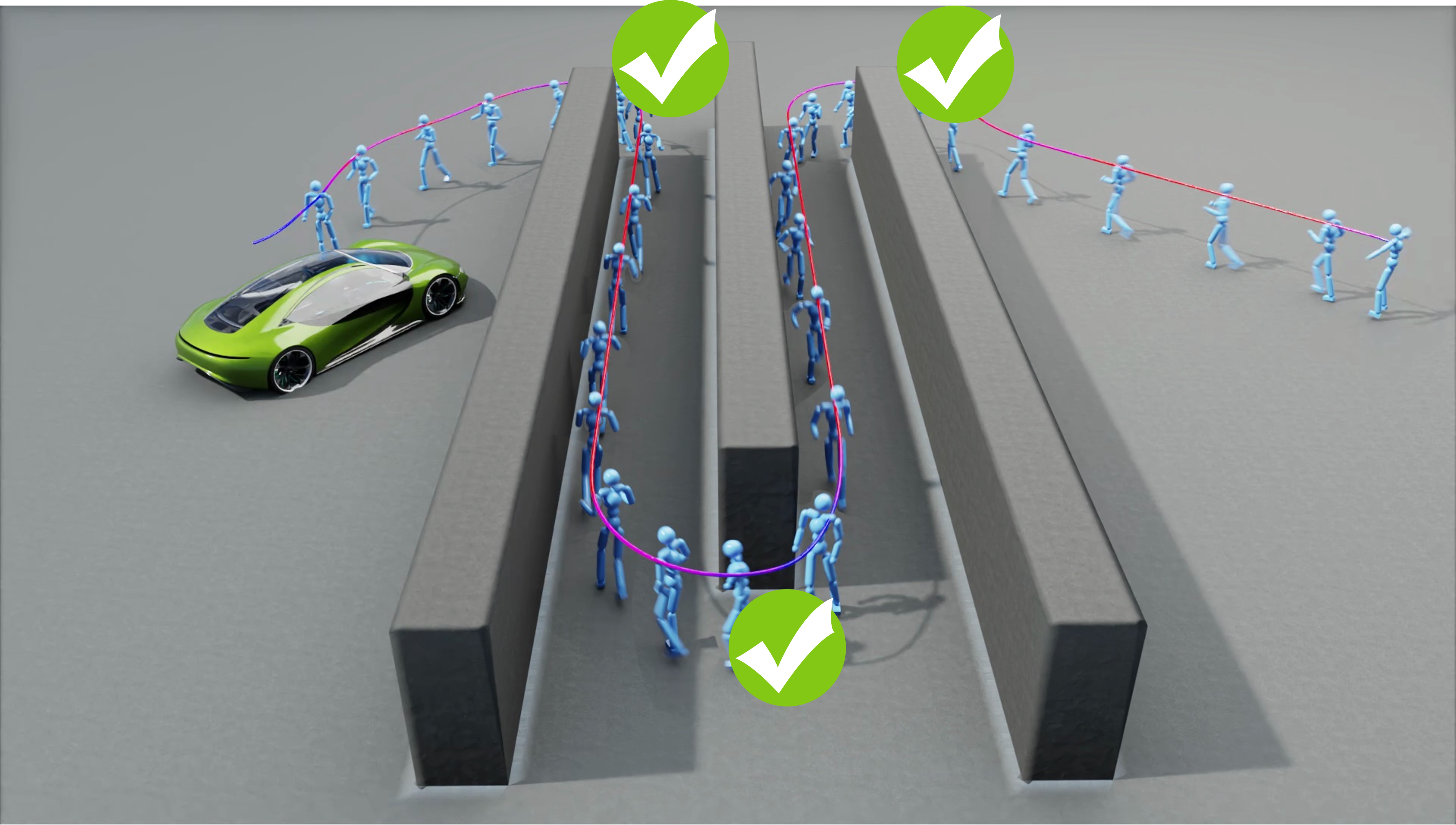}
         \caption{\textbf{Slalom test with adaptive speed (ours).} The humanoid follows the path well, never hitting the walls.}
         \label{fig: slalom dynamic}
     \end{subfigure}
     \caption{\textbf{Comparison of consistent and adaptive speed profiles on the slalom test.} The colors represent the speed, with red indicating speeds above 3.5m/s and blue below 1m/s.}
    \label{fig: slalom}
\end{figure*}

% \newpage
% \onecolumn
% \newpage
%\twocolumn

% Appendix
\clearpage
\appendix

\section{Data curation}
To train our motion controller, we consider three datasets. First, the entire AMASS dataset contains various motions, such as dancing, cartwheels, and sitting. Second, we consider an automatic curation method. This uses a large language model (LLM, ChatGPT4 \cite{openai2023gpt}). We the LLM with the textual descriptions of each movement and ask it to categorize the motions into five classes: [`walking', `running', `crouching', `crawling', and others]. These textual labels are extracted from the HumanML3D dataset \cite{guo2022generating}. This filtering reduced the size of the dataset down to 600 motions. Finally, we also tested a manual curation procedure. For manual curation, we handpick 75 motion captures, roughly 20 from each desired motion type, and create a small, tailor-made dataset.

\section{Additional results}
In Tables \ref{tab: locomotion controller a} - \ref{tab: locomotion controller e} we present the full results for the ablation study given in Table \ref{tab: locomotion ablation}. We show the XY and Z errors for every choice of data filtering and path generator speed samples, for every terrain and every motion type. 

\begin{table*}[ht]
    \centering
    \caption{\textbf{No data filtering. Naive path generator. Locomotion controller errors.}}
    \begin{tabular}{l|c|c|c|c|c|c|c|c}
                    & \multicolumn{2}{c|}{\textbf{Crawl }} & \multicolumn{2}{c|}{\textbf{Crouch-walk}} & \multicolumn{2}{c|}{\textbf{Walk}} & \multicolumn{2}{c}{\textbf{Run}}  \\ \toprule 
                   \textbf{Terrain} & XY-err $\left[m\right]$ & Z-err$\left[m\right]$ & XY-err $\left[m\right]$& Z-err $\left[m\right]$& XY-err$\left[m\right]$ & Z-err $\left[m\right]$& XY-err$\left[m\right]$ & Z-err$\left[m\right]$ \\ \hline
        Flat    & 1.11 & 0.20 & 0.43 & 0.11 & 0.35 & 0.31 & 0.76 & 0.36 \\ \hline
        Slope   & 1.12 & 0.20 & 0.46 & 0.11 & 0.38 & 0.32 & 0.82 & 0.36 \\ \hline
        Rough   & 1.39 & 0.19 & 0.53 & 0.13 & 0.44 & 0.34 & 1.06 & 0.40 \\ \hline
        Stairs  & 1.44 & 0.21 & 0.78 & 0.16 & 0.53 & 0.36 & 1.19 & 0.42
    \end{tabular}
    \label{tab: locomotion controller a}
\end{table*}

\begin{table*}[ht]
    \centering
    \caption{\textbf{No data filtering. Coupled path generator. Locomotion controller errors.}}
    \begin{tabular}{l|c|c|c|c|c|c|c|c}
                    & \multicolumn{2}{c|}{\textbf{Crawl }} & \multicolumn{2}{c|}{\textbf{Crouch-walk}} & \multicolumn{2}{c|}{\textbf{Walk}} & \multicolumn{2}{c}{\textbf{Run}}  \\ \toprule 
                   \textbf{Terrain} & XY-err $\left[m\right]$ & Z-err$\left[m\right]$ & XY-err $\left[m\right]$& Z-err $\left[m\right]$& XY-err$\left[m\right]$ & Z-err $\left[m\right]$& XY-err$\left[m\right]$ & Z-err$\left[m\right]$ \\ \hline
        Flat    & 0.43 & 0.07 & 0.39 & 0.07 & 0.37 & 0.31 & 0.74 & 0.36 \\ \hline
        Slope   & 0.49 & 0.08 & 0.42 & 0.07 & 0.40 & 0.32 & 0.81 & 0.38 \\ \hline
        Rough   & 0.66 & 0.09 & 0.49 & 0.08 & 0.45 & 0.34 & 1.03 & 0.42 \\ \hline
        Stairs  & 0.94 & 0.13 & 0.67 & 0.10 & 0.53 & 0.35 & 1.13 & 0.43
    \end{tabular}
    \label{tab: locomotion controller b}
\end{table*}

\begin{table*}[ht]
    \centering
    \caption{\textbf{Naive data filtering. Naive path generator. Locomotion controller errors.}}
    \begin{tabular}{l|c|c|c|c|c|c|c|c}
                    & \multicolumn{2}{c|}{\textbf{Crawl }} & \multicolumn{2}{c|}{\textbf{Crouch-walk}} & \multicolumn{2}{c|}{\textbf{Walk}} & \multicolumn{2}{c}{\textbf{Run}}  \\ \toprule 
                   \textbf{Terrain} & XY-err $\left[m\right]$ & Z-err$\left[m\right]$ & XY-err $\left[m\right]$& Z-err $\left[m\right]$& XY-err$\left[m\right]$ & Z-err $\left[m\right]$& XY-err$\left[m\right]$ & Z-err$\left[m\right]$ \\ \hline
        Flat    & 0.62 & 0.15 & 0.26 & 0.05 & 0.29 & 0.27 & 0.60 & 0.30 \\ \hline
        % Slope   & 0.49 & 0.08 & 0.42 & 0.07 & 0.40 & 0.32 & 0.81 & 0.38 \\ \hline
        Rough   & 1.16 & 0.17 & 0.37 & 0.07 & 0.35 & 0.28 & 0.87 & 0.36 \\ \hline
        Stairs  & 1.24 & 0.19 & 0.52 & 0.11 & 0.43 & 0.30 & 1.02 & 0.37
    \end{tabular}
    \label{tab: locomotion controller c}
\end{table*}

\begin{table*}[ht]
    \centering
    \caption{\textbf{Naive data filtering. Coupled path generator. Locomotion controller errors.}}
    \begin{tabular}{l|c|c|c|c|c|c|c|c}
                    & \multicolumn{2}{c|}{\textbf{Crawl }} & \multicolumn{2}{c|}{\textbf{Crouch-walk}} & \multicolumn{2}{c|}{\textbf{Walk}} & \multicolumn{2}{c}{\textbf{Run}}  \\ \toprule 
                   \textbf{Terrain} & XY-err $\left[m\right]$ & Z-err$\left[m\right]$ & XY-err $\left[m\right]$& Z-err $\left[m\right]$& XY-err$\left[m\right]$ & Z-err $\left[m\right]$& XY-err$\left[m\right]$ & Z-err$\left[m\right]$ \\ \hline
        Flat    & 0.67 & 0.06 & 0.25 & 0.06 & 0.25 & 0.23 & 0.57 & 0.29 \\ \hline
        Slope   & 0.55 & 0.07 & 0.26 & 0.07 & 0.30 & 0.23 & 0.64 & 0.30 \\ \hline
        Rough   & 0.78 & 0.09 & 0.32 & 0.07 & 0.34 & 0.26 & 0.85 & 0.35 \\ \hline
        Stairs  & 0.80 & 0.13 & 0.48 & 0.10 & 0.42 & 0.28 & 1.01 & 0.36
    \end{tabular}
    \label{tab: locomotion controller d}
\end{table*}

\begin{table*}[ht]
    \centering
    \caption{\textbf{Curated data filtering. Naive path generator. Locomotion controller errors.}}
    \begin{tabular}{l|c|c|c|c|c|c|c|c}
                    & \multicolumn{2}{c|}{\textbf{Crawl }} & \multicolumn{2}{c|}{\textbf{Crouch-walk}} & \multicolumn{2}{c|}{\textbf{Walk}} & \multicolumn{2}{c}{\textbf{Run}}  \\ \toprule 
                   \textbf{Terrain} & XY-err $\left[m\right]$ & Z-err$\left[m\right]$ & XY-err $\left[m\right]$& Z-err $\left[m\right]$& XY-err$\left[m\right]$ & Z-err $\left[m\right]$& XY-err$\left[m\right]$ & Z-err$\left[m\right]$ \\ \hline
        Flat    & 0.29 & 0.09 & 0.25 & 0.04 & 0.21 & 0.21 & 0.49 & 0.27 \\ \hline
        Slope   & 0.34 & 0.12 & 0.26 & 0.04 & 0.24 & 0.22 & 0.57 & 0.28 \\ \hline
        Rough   & 0.50 & 0.13 & 0.29 & 0.05 & 0.27 & 0.23 & 0.78 & 0.32 \\ \hline
        Stairs  & 0.81 & 0.17 & 0.43 & 0.07 & 0.36 & 0.26 & 0.97 & 0.35
    \end{tabular}
    \label{tab: locomotion controller e}
\end{table*}

\section{Additional technical details for Section~\ref{sec:planner}}
As a final step, the output format of the path planner (x,y,z,v) is translated into a format suitable for the motion controller. This controller requires $(x,y,z)$ waypoints spaced at fixed time intervals, where a larger waypoint spacing corresponds to a higher speed. The conversion from the spatially defined path to this temporal trajectory format is accomplished through an iterative algorithm supplemented by linear interpolation. 

\section{Additional experiments for the speed controller}
\begin{figure}[ht]
    \centering
    \includegraphics[width=0.8\linewidth]{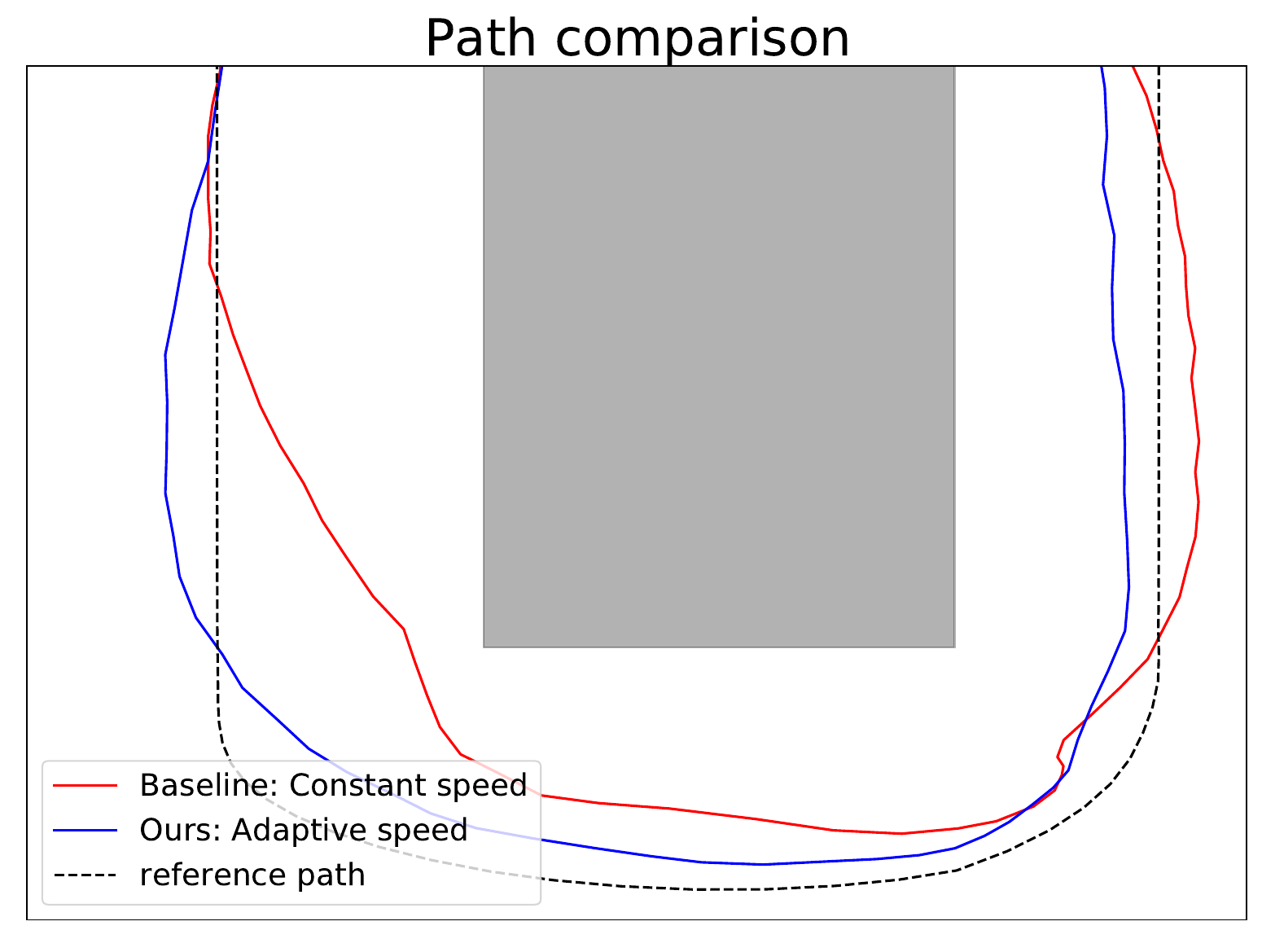}
    \caption{\textbf{Comparison of planned paths} (blue) Our path with adaptive speed. (red) A path with a constant speed that has the same average speed as the adaptive-speed path. (dashed black) the desired reference path. %The constant speed was selected so that the two runs have exactly the same completion time. 
    When using a constant speed, the humanoid hits the wall at its corners. Using the adaptive speed generates a fast but smoother path, that is far from the obstacle but still follows follows well the desired path better.}
    \label{fig: slalom curve}
\end{figure}

Here, we provide additional results on the slalom test from Figure~\ref{fig: slalom}. We record the paths the humanoid traversed with a constant speed, as well as our speed controller output, and compare them with the desired reference path. The constant speed was selected so the two runs had the same completion time. The result is shown in Figure~\ref{fig: slalom curve}. Despite having the same average speeds, in the constant-speed run, the humanoid has a difficult time matching the speed during the steep turns and hits the wall at its corners. This does not happen with the adaptive speed run, which follows the desired path better.

% In Figure~\ref{fig: slalom pareto 2}, we report the average 3D disposition error on the y-axis versus the average completion time. As seen, our planner achieved the Pareto front.
% \begin{figure}[ht]
%     \centering
%     \includegraphics[width=0.87\linewidth]{figures/planner/slalom_pareto2.pdf}
%     \caption{\textbf{Disposition error and Completion Time: Pareto curve in the Slalom test.} Testing our our speed controller vs. constant speed. Our planner obtains lower disposition error while retaining faster completion.}
%     \label{fig: slalom pareto 2}
% \end{figure}

\section{Supplementary files}
We provide as addition to the paper supplementary videos to demonstrate PlaMo:
\begin{enumerate}
    \item Videos of the locomotion controller in terrain X motion type that correspond to Table \ref{tab: locomotion controller} and Figure \ref{fig: locomotion example}. 
    \item Videos of the humanoid doing the Slalom scene presented in Figure \ref{fig: slalom}.
    \item Videos of randomized route and motion types in a diverse scene, corresponding to Figure \ref{fig: playground}
\end{enumerate}
We also provide yaml files that correspond to the filtered dataset considered for the ablation study given in Table \ref{tab: locomotion ablation}.

% \section{State and action space}
% The actions from the policy specify target positions for PD controllers positioned at each of the character's joints, which in turn produce control forces that drive the motion of the character.

% % TODO
% \section{Simulation details}\label{appendix:sim_details}

% \section{Supplementary material description}

% TODO -- a less researchy discussion on design choices and some thoughts

\end{document}